\documentclass[sigplan,screen]{acmart}
\acmSubmissionID{4360}

\usepackage{natbib}
\setcitestyle{numbers,square}
\usepackage{amsmath}
\usepackage{cuted}
\usepackage{algorithm}
\usepackage{algorithmicx}
\usepackage{algpseudocode}
\usepackage{flushend}
\usepackage{enumerate}

\renewcommand\footnotetextcopyrightpermission[1]{}

\makeatletter
\renewcommand{\maketag@@@}[1]{\hbox{\m@th\normalsize\normalfont#1}}%
\makeatother

\AtBeginDocument{%
  }

\begin{document}

\raggedbottom

\title{ITCMA: A Generative Agent Based on a Computational Consciousness Structure}

\author{Hanzhong Zhang}
\affiliation{%
 \institution{Kunming University of Science and Technology}
     \city{Kunming}
 \state{Yunnan Province}
 \country{China}}
\email{armihia@foxmail.com}

\author{Jibin Yin}
\affiliation{%
 \institution{Kunming University of Science and Technology}
     \city{Kunming}
 \state{Yunnan Province}
 \country{China}}
\email{yjblovelh@aliyun.com}

\author{Haoyang Wang}
\affiliation{%
 \institution{Kunming University of Science and Technology}
     \city{Kunming}
 \state{Yunnan Province}
 \country{China}}
\email{HaoyangWang123@outlook.com}

\author{Ziwei Xiang}
\affiliation{%
 \institution{Huaqiao University}
     \city{Xiamen}
 \state{Fujian Province}
 \country{China}}
\email{zjcutomunist@gmail.com}

\renewcommand{\shortauthors}{Zhang et al.}
\acmConference[UIST'24]{ACM Conference}{Oct 2024}{Pittsburgh, PA, USA}%

\begin{abstract}
Large Language Models (LLMs) still face challenges in tasks requiring understanding implicit instructions and applying common-sense knowledge. In such scenarios, LLMs may require multiple attempts to achieve human-level performance, potentially leading to inaccurate responses or inferences in practical environments, affecting their long-term consistency and behavior. This paper introduces the Internal Time-Consciousness Machine (ITCM), a computational consciousness structure to simulate the process of human consciousness. We further propose the ITCM-based Agent (ITCMA), which supports action generation and reasoning in open-world settings, and can independently complete tasks. ITCMA enhances LLMs' ability to understand implicit instructions and apply common-sense knowledge by considering agents' interaction and reasoning with the environment. Evaluations in the Alfworld environment show that trained ITCMA outperforms the state-of-the-art (SOTA) by 9\% on the seen set. Even untrained ITCMA achieves a 96\% task completion rate on the seen set, 5\% higher than SOTA, indicating its superiority over traditional intelligent agents in utility and generalization. In real-world tasks with quadruped robots, the untrained ITCMA achieves an 85\% task completion rate, which is close to its performance in the unseen set, demonstrating its comparable utility and universality in real-world settings. 
\end{abstract}



\maketitle

\section{Introduction}

With the advancement of artificial intelligence (AI), the term “agent” has been increasingly used to describe entities that demonstrate intelligent behavior and possess qualities such as autonomy, reactivity, pro-activeness, and social ability \cite{wooldridge1995intelligent, goodwin1995formalizing}. The development of large language models (LLMs) has provided new research perspectives on the use of agents. Park et al. \cite{park2023generative} introduced generative agents—intelligent entities utilizing generative LLMs to simulate believable human behavior—and demonstrated their ability to produce credible individual and emergent collective behavior in simulations. Generative agents are capable of making various inferences about themselves, other agents, and their environment; they can formulate daily plans reflecting their characteristics and experiences, execute these plans, react, and revise plans as needed.

The foundation of such ability is a novel agent architecture that combines LLMs with mechanisms for synthesizing and retrieving relevant information, thus providing conditions for the output of language models. Without these mechanisms, LLMs could still output behavior for agents, but the agents might not react correctly based on past experiences, fail to make crucial inferences, or struggle to maintain long-term consistency \cite{park2023generative}.

As the interaction cycles of LLM-based agents grow, two issues arise with this structure. The first issue relates to the length of the historical record. LLM-based agents process previous interactions in natural language format, appending the historical record to each subsequent input. As these records extend, they might exceed the constraints of the Transformer architectures most LLM-based agents rely on. In such cases, the system might truncate certain contents that were input into the LLM in the early stages. The second issue is the difficulty of extracting relevant memories. As agents accumulate a large volume of historical observations and action sequences, their memory burden increases continuously. This makes establishing connections between relevant topics increasingly challenging, potentially resulting in the agent's responses being inconsistent with the ongoing context \cite{xi2023rise}.

Due to the current limitations of LLMs acting as the "brain," recently research on LLM-based agents focuses on augmenting them with an additional layer of structures, such as LangChain \cite{Pandya2023AutomatingCS}, to optimize them. However, although these studies demonstrate the potential to better utilize LLMs through structural modifications, and some even incorporate modeling of human cognitive processes \cite{Toy2024MetacognitionIA}, the emergence of processes similar to human consciousness remains largely unexplored in existing research. This is mainly because these studies typically focus on solving specific tasks—tasks that are often specially defined and require relevant datasets to train the agents. To some extent, this sacrifices the excellent explore and exploit (EE) capabilities of generative agents, as their utility relies on the training outcomes within specific environments.

To enable generative agents to possess sufficient generalization capabilities and adapt to new environments without prior training, we believe the key lies in simulating a human-like consciousness process. This is distinct from the cognitive processes commonly discussed in the field of artificial intelligence. An agent with a similar consciousness process is not modeled as a "rational economic agent" but rather as a "bounded rationality" model. This means that, an agent makes decisions by constructing and testing a relatively small number of possible strategies, then selecting behaviors that sufficiently balance intuition and efficacy \cite{Simon1989}.

The most typical work here comes from cognitive science and philosophy of mind, a branch of them pays special attention to the problem of consciousness. Researchers in it discuss the problem of consciousness and human intelligence and often draw inspiration from phenomenology, especially Merleau-Ponty's phenomenology of perception \cite{Gallagher01, Thompson2010,Zlatev2008}. Interdisciplinary research in this area has revealed that intelligence relies on consciousness processes to exist. That is, intelligence is not the stimulus-response mode claimed by behaviorism (represented by reinforcement learning in AI research), nor is it an "emergence from neuroelectricity" claimed by connectionism (represented by deep learning in artificial intelligence research). An expression of mechanical determinism is not applicable in the study of intelligent structures \cite{Merleau-Ponty2012,thompson2001radical}.

Therefore, we believe that, in order to design a more effective generative agent architecture, it is necessary to consider a framework that simulates the human consciousness process. In this paper, we propose a computational consciousness structure for generative agents, called the Internal Time-Consciousness Machine (ITCM). It is particularly worth mentioning that the computational structure of consciousness here does not equate to generating a structure equivalent to human consciousness in a computer. Instead, it refers to a simple and abstract model, like the Conscious Turing Machine \cite{blum2022theory}, which aims to algorithmically simulate the observed processes of consciousness. The basic unit of ITCM, the field, encompasses the spatial topology of perceived objects. The ITCM consists of three modules: (1) a consciousness channel module that constructs a temporal prediction process for the field to obtain a possible future change in the environment after action, (2) a memory module that aids in understanding the current environment, and (3) a driving force module that provides selection weights for the action space based on computations from the other modules. Subsequently, we designed a generative agent architecture based on ITCM, called the ITCM-based agent (ITCMA). This architecture aims to optimize the mechanism by which generative agents generate behavioral outputs based on their environment and experiences, and enhance the EE and generalization abilities of generative agents. Our main contributions are as follows:

\begin{itemize}
\item We propose the ITCM as a reconstruction of the underlying architecture of generative agents. This computational consciousness structure can assist generative agents in handling complex tasks with greater flexibility and intelligence while enhancing their interpretability, making their behaviors easier to understand and predict.

\item Technically, based on the proposed ITCM structure, we introduce an ITCMA and validate its effectiveness in both of the real life scenario environments described by text, as well as in real-world robotic environments. Based on the ITCM structure, the ITCMA demonstrated strong EE and generalization abilities in experimental environments, achieving high utility even without prior training in these environments.
\end{itemize}

\section{Related Work}

\subsection{LLMs as agents}
The emergence of LLMs has brought new avenues for intelligent agents. In recent years, there has been a proliferation of LLM-based agent architectures \cite{wang2022self, sumers2023cognitive, xi2023rise}. These architectures primarily focus on two aspects: planning and tools \cite{Liu2024FromLT}.

The most prominent research focus in the planning domain is Chain of Thought (CoT) reasoning. This involves eliciting logical reasoning from LLMs through CoT prompts. Initially proposed by Wei et al. \cite{wei2022chain}, CoT's improvement lies in presenting a series of reasoning steps (manually constructed) for the answer part of examples before providing the final answer. The logic is to teach the model to output reasoning steps gradually and then deliver the result.

Building upon this, Wang et al. \cite{wang2023survey} introduced a novel decoding strategy—self-consistency, to replace the naive greedy decoding used in CoT prompts. This strategy capitalizes on the intuition that complex reasoning problems often allow for multiple different ways of thinking, leading to unique correct answers. Zhou et al. \cite{zhou2022least} proposed a novel prompting strategy, from minimal to maximal prompts. The key idea of this strategy is to decompose complex problems into a series of simpler sub-problems, which are then solved sequentially. The answers to previously solved sub-problems aid in solving each subsequent sub-problem.

Yao et al. \cite{yao2022react} introduced the ReAct framework to generate reasoning trajectories and actions for specific tasks in an interleaved manner, enhancing the synergy between the two: reasoning trajectories aid the model in inducing, tracking, and updating action plans, and handling anomalies, while actions enable it to interface with external sources (such as knowledge bases or environments) to gather more information. Building on this, Liu et al. \cite{Liu2024FromLT} proposed the RAISE architecture, which integrates a dual-component memory system reflecting human short-term and long-term memory to maintain context and continuity in dialogue. This approach enhances the controllability and adaptability of agents in complex, multi-turn conversations.

In the tools domain, the primary focus is on the LLMs' ability to leverage external tools and resources. Numerous studies have demonstrated the effectiveness of LLMs using external tools and APIs. Schick et al. \cite{schick2024toolformer} proposed Toolformer, which can determine which APIs to call, when to call them, what parameters to pass, and how to best integrate the results into future token predictions. Shen et al. \cite{shen2024hugginggpt} introduced HuggingGPT, an LLM-driven agent that leverages LLMs to connect various artificial intelligence models within the machine learning community to solve AI tasks.

In some more intriguing domains, conversational agents have also received widespread discussion. Shao et al. \cite{shao2023character} proposed Character-LLM, which enables LLMs to embody specific characters by training models with profiles edited to represent the experiences of those characters. Chen et al. \cite{chen2023chatcot} introduced ChatCoT, which models CoT reasoning as multi-turn conversations to use tools more naturally through chatting. Chae et al. \cite{Chae2023DialogueCD} proposed a knowledge extraction framework to facilitate dialogue CoT reasoning within the conversation context and subsequently introduced a Dialogue Reasoning Chain DOCTOR to provide reliable CoT reasoning for response generation.

Apart from constructing LLM-based agents via prompts, some research focuses on fine-tuning methods. Zeng et al. \cite{zeng2022glm} introduced AgentTuning for fine-tuning the Llama 2 model to produce AgentLM. Chen et al. \cite{chen2023fireact} proposed FireAct, a new method for fine-tuning language models using trajectories and prompting across multiple tasks, demonstrating that having a more diverse set of fine-tuning data can further improve agent performance.

However, as previously mentioned, current research focuses on improving the training effectiveness of generative agents on specific tasks through enhancements to planning and tool modules. This sacrifices some of the superiority of generative agent structures in EE performance. In this paper, we aim to enhance the EE and generalization abilities of generative agents by proposing the Internal Time-Consciousness Machine (ITCM), a computational consciousness structure adapted for generative agents. ITCM simulates human consciousness processes rather than cognitive processes. This is achieved by reconstructing the underlying logic of generative agents, thereby improving their EE and generalization abilities.

\subsection{Computational Consciousness Structure}

\subsubsection{Integrated Information Theory}

The central claim of integrated information theory (IIT) is that a physical system is conscious when and only when it is the maximum of the integrated information $\phi$ \cite{oizumi2014phenomenology, tononi2016integrated}. As defined by the IIT, integrated information can be roughly described as a measure of the extent to which a system is causally bound to its own past and future states and the dependence of these constraints on the causal interconnectedness between system components. If a system has more integrated information than any overlapping system (e.g., a smaller system that is a part of it or any larger system containing it), then this system has the maximum comprehensive information. Consciousness needs the largest $\phi$, not just a non-zero value of $\phi$ (this is called the exclusion theorem). The IIT also claims that consciousness is identical to the maximum $\phi$.

The IIT has been met with some philosophical objections \cite{morch2019consciousness}. One of these objections is that consciousness intuitively looks like an intrinsic property, but maximal $\phi$ is an extrinsic property. Therefore, if this intuition is correct, then consciousness cannot be the same as maximal $\phi$ \cite{Chalmers2016}.

\subsubsection{Consciousness Turing Machine}

Influenced by Turing machines (TMs) and conscious global workspace theory (GWT), Blum \cite{blum2021theoretical} combined computational complexity theory and machine learning knowledge to propose a formal theoretical computer model called a consciousness Turing machine (CTM).

According to Blum \cite{blum2022theory}, CTM is not meant to model the brain or imply neural correlates of consciousness. It is a simple abstract model of consciousness aimed at understanding consciousness and its related phenomena. Therefore, the definition of CTM adopts a theoretical computer science perspective, forming a conscious Global Workspace Theory (GWT) mathematically. 

Therefore,the basis of consciousness in the CTM uses Baar’s theater of consciousness hypothesis, which likens consciousness to the performance of dramatic actors on a stage of working memory. Their performance occurs under the observation of a group of spectators sitting in the dark (unconscious processors). The stage of the GWT is represented by short-term memory (STM) that contains the content of the CTM consciousness at any moment. The spectators are represented by powerful processors. Each processor has its own expertise, and these processors make up the CTM’s long-term memory (LTM). These processors make predictions and receive feedback from the CTM’s world. The learning algorithms inside each processor improve the processor’s behavior based on this feedback.

When all LTM processors receive these contents via this broadcast, CTM becomes consciously aware of these contents. In CTM, the definition of consciousness roughly aligns with what cognitive neuroscientists refer to as "attention." The feeling of consciousness in CTM corresponds roughly to what cognitive neuroscientists refer to as "consciousness" or "subjective consciousness."

\section{The Internal Time-Consciousness Machine based Agent (ITCMA)}

It is undeniable that the IIT and CTM suggest ways to build algorithms based on different aspects of consciousness research. However, the question remains as to whether there is truly a computation-based theory of consciousness, even though there are many similarities between the brain and the structure of the digital computer. Thagard \cite{Thagard2005} summarizes seven challenges to the study of representation computation, and the asymmetry between cognition and computation occupies the first position in this list of challenges. In Appendix A, We discuss in detail the necessity of focusing on constructing a pseudo-consciousness model corresponding to intelligence by turning our attention to first-person phenomenology. However, at present, we need to first introduce the overall framework of the internal time-consciousness machine based agent (ITCMA) as a generative agent structure. As for how the internal time-consciousness machine (ITCM), which is the computational consciousness structure behind ITCMA, is proposed in accordance with phenomenological thinking, we will discuss it in Appendix B. In the following part of this section, we will mainly discuss the model form of ITCM.

\subsection{The overall structure of ITCMA}

Park et al. \cite{park2023generative} proposed the original generative agent structure in which all content is recorded and inferred in natural language, enabling the structure to utilize an LLM. Influenced by this, Xi et al. \cite{xi2023rise} proposed a general conceptual framework for generative agents that includes three key components: the brain, perception, and action.

The structure of the ITCMA is similar in that perception serves as input from the environment, and action serves as output to the environment. After receiving action, the environment forces the agent to form a perception as a re-action to it. Reaction is the carrier of this action—the brain is not equivalent to the LLM here but is replaced by the complete ITCMA, and the LLM is only used as a tool. The structure is shown in Figure 1.

\begin{figure}[h]
  \centering
  \includegraphics[width=\linewidth]{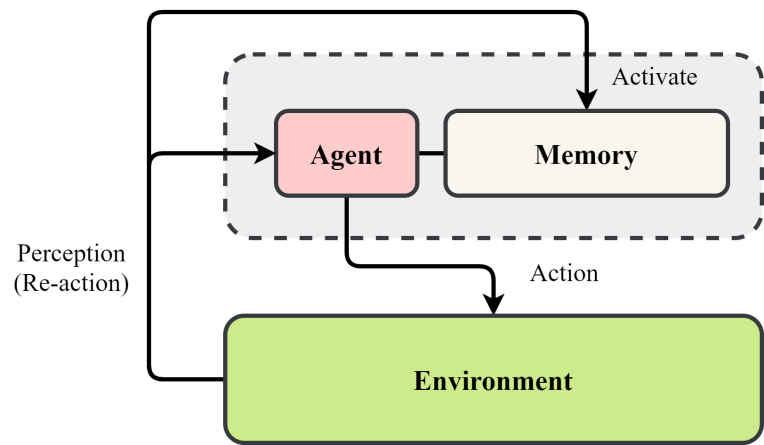}
  \caption{Conceptual framework of ITCMA.}
\end{figure}

Figure 1 is based on the specific ITCMA structure, which can be further refined. Based on the foundational structure of the brain, perception, and action, we first define the basic perceptual unit of ITCMA, the field, which includes the spatial topology of perceived objects. Following this, we define three modules for ITCMA:

\begin{enumerate}[1.]
\item \textbf{The consciousness channel module}, which constructs the temporal prediction process for the field.
\item \textbf{The memory module}, which aids in understanding the current environment.
\item \textbf{The driving force module}, which provides selection weights for the action space based on computations from the other modules.
\end{enumerate}

The interaction between these three modules follows the following process: after receiving environmental information, the consciousness channel module first changes some of its own content (retention and primal impression), and inputs this part to the memory module to obtain the activated memory. This memory will be used to generate the remaining part of the consciousness channel module (Protention). After that, the contents of these two modules will be input to the driving force module to calculate the weight of the action space, and then the LLM will make a decision on the action.

In more detailed expressions, at time $t$, the current perception provided by environment (primal impression) $PI^t$ and the list retention $Re^t$ made up of perceptions from previous moments are combined. The ITCMA uses this to extract the activated memory $AM$ and its subsequent content from long-term memory. These three together form consciousness channel $C^t$, at time $t$ after receiving $C^{t-1}$ of time $t-1$. Through $C^t$, ITCMA can use a time series forecasting model (TSFM) to calculate for each action in the action space, in order to obtain the possible changes in the environment (Protention) $Pro^t$  after they are executed. It will then be used to calculate the selection weight driving force $d^t$ for each action. The above process is ultimately provided to the LLM in natural language to select an action to execute. The entire process is shown in Figure 2.

\begin{figure}[h]
  \centering
  \includegraphics[width=\linewidth]{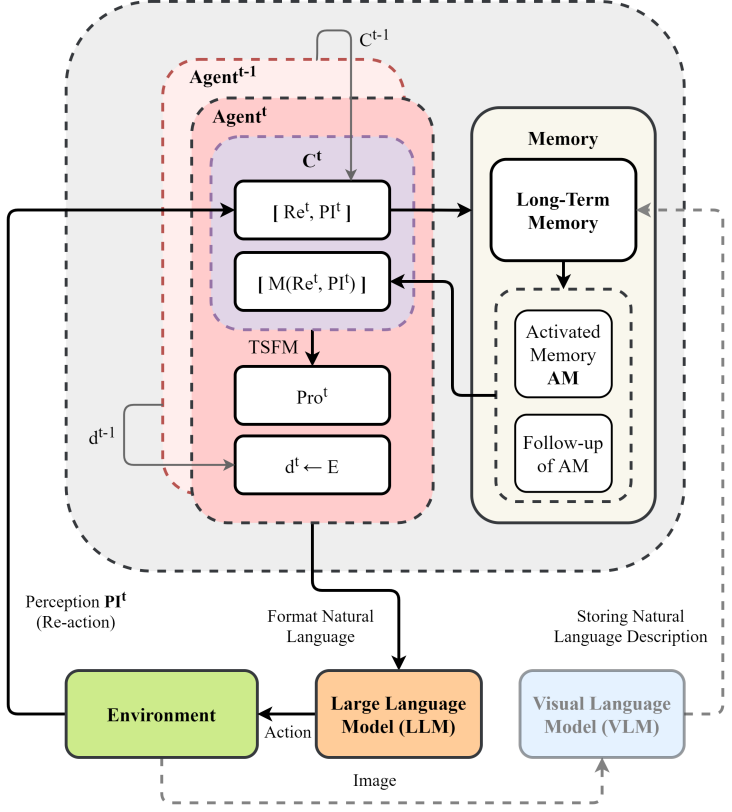}
  \caption{Detailed structure of the ITCMA.}
\end{figure}

Even though LLM is not used as a direct brain like in other generative agent structures, its role in ITCMA is also very central. Its role is to understand the information provided by the perception module and, based on this information, combine its own understanding of the world to output an action that the agent will execute. Therefore, the working process of ITCMA can be compared to a few-shot CoT that mimics the human consciousness thinking process. The ITCM structure can give LLM a regular human-like thinking logic through an explore and exploit (EE) process and a context model centered on the extracted more suitable environmental memory, thereby establishing a new rule system beyond LLM.

In addition, due to the ITCM constructing an agent that simulates consciousness without experience, using a visual language model (VLM) to aid in modeling is an option here. In complex task environments, the information captured by vision is not only the spatial topology of an object but also includes a holistic understanding of it. Using the VLM here to characterize the visual image in natural language form and add it to the ITCMA workflow can somewhat bypass its early process of understanding of spatial relationships.

However, due to the enhancement of the VLM’s unique visual ability, its language generation ability is slightly insufficient compared to an LLM of the same scale. Therefore, it is only used as a descriptive tool here and cannot replace the original LLM used to generate actions. In an evaluation based on real-world quadrupedal robot environments, we will use the VLM to accomplish an understanding of three-dimensional spatial information. 

In the next section, we will introduce the details and mechanisms of each module. Appendix C includes the specific implementation details and examples of the ITCMA.

\subsection{Modules in ITCM}

\subsubsection{Field}

A "field" is a functional and exclusive activity location of conscious experience, as identified by Husserl \cite{Husserl2008}. According to Merleau-Ponty \cite{Merleau-Ponty2012}, the perceptual subject is always in the midst of other objects; it is always a part of the field. Without a mechanism such as the perception field, our perception would not be continuous at all. Therefore, let us imagine a spherical coordinate system: the perceptual subject (that is, the agent) is always in the middle of the sphere and facing forward, and the object that can be perceived is described by the set of spherical coordinates ($\theta$, $\varphi$, $\gamma$). When we obtain these sets of represented objects, the spherical coordinate system that can be transformed by the agent's actions becomes the agent's phenomenal field. 

A field is therefore decomposed into two parts: a perceptual field and a phenomenal field. Among them, the phenomenal field is a spherical coordinate system centered on the subject, which contains many objects, and it can make some predictive changes to these objects according to experience; the perceptual field is a part of the phenomenal field, which is includes those objects that are directly perceived. The objects contained by it can be forcibly updated by a perceived "fact." For example, during a speech, the speakers can directly perceive the audience's reactions and also consider, through imagination, the possible discussions among audience members that they cannot directly hear. A diagram of the phenomenal and perceptual fields is shown in Figure 3.

\begin{figure}[h]
  \centering
  \includegraphics[width=\linewidth]{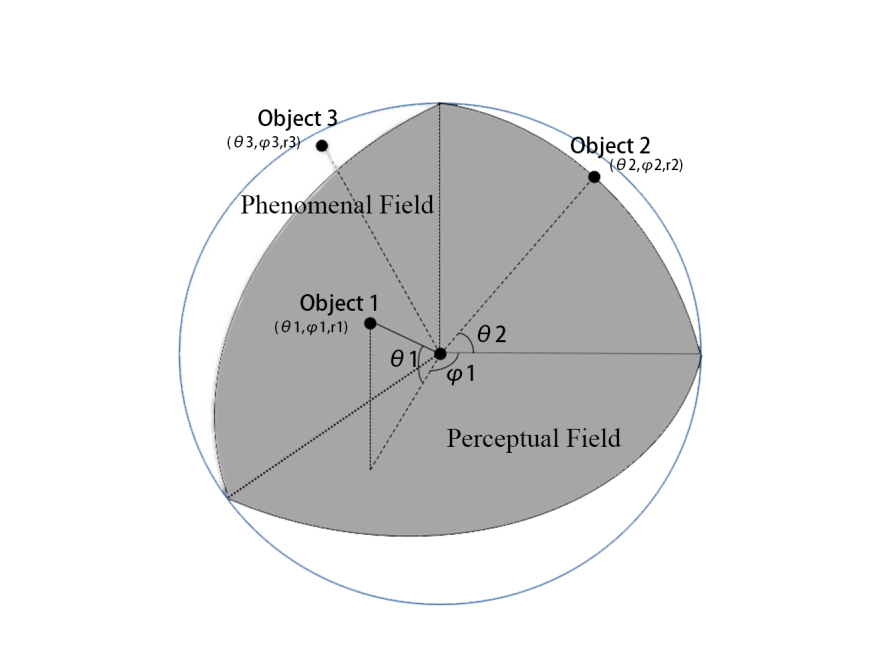}
  \caption{Phenomenal and perceptual fields. The white area represents the phenomenal field, within which object 3 has not been perceived by the agent. The gray area represents the perceptual field, within which objects 1 and 2 are directly perceived by the agent.}
\end{figure}

In this illustration, the phenomenal field is precisely such a spherical coordinate: it is centered on the agent, and scoped by the agent’s perceptual domain that decays with space. If there are perceived objects, they contain two attributes: an attribute $N$ representing their own content and an attribute $pos \leftarrow \begin{bmatrix}  \theta & \varphi & \gamma \end{bmatrix}$ representing their own position. For $N$, we can easily associate word vectors. A transformation is needed to expand $N$ into a specific -dimensional vector that is sufficient to calculate the distance between object meanings. Therefore, for a field containing $m$ objects, it can be unfolded in the following form after training:

\begin{equation}
f=
\begin{bmatrix}
 N_{1} & \theta_{1} & \varphi_{1} & \gamma_{1}\\
 \vdots & \vdots & \vdots & \vdots\\
 N_{m} & \theta_{m} & \varphi_{m} & \gamma_{m}\\
 \end{bmatrix}
 =
\begin{bmatrix}
 N_{1}^1 & \dots & N_{1}^n & \theta_{1} & \varphi_{1} & \gamma_{1}\\
 \vdots & \ddots & \vdots & \vdots & \vdots & \vdots\\
 N_{m}^1 & \dots & N_{m}^n & \theta_{m} & \varphi_{m} & \gamma_{m}\\
 \end{bmatrix}
\end{equation}

\subsubsection{Consciousness Channel Module}

The field provides a topological modeling of the perceived objects. However, consciousness is a continuous process. We can think of one $f$-representation as a sampling of the originally continuous consciousness at time $t$. By establishing a continuous consciousness stream channel through multiple frames, we can construct a temporal structure that will be used for future predictions.

According to Husserl \cite{Husserl2008}, the basic unit of temporality is a temporal field that comprises all three temporal modes of present, past, and future. The modeling of the past, we refer to as (1) \textbf{Retention}. It consists of several temporally continuous fields. It should be emphasized that retention should be distinguished from recall. There is a significant difference between a person’s impression of something that just happened (retention) and a person’s memory of a past event. A retentionis directly connected to a (2) \textbf{Primal impression}. Primal impression is a single field that represents the raw content of information perceived at the present moment. By combining the temporal elements of retention and primal impression, we can predict how the field will change in the next moment, thus obtaining a direction towards future information. The possible change state of the field at the next moment after the execution of the action is called (3) \textbf{Protention}. It should be emphasized that protention is not equivalent to a primal impression at time $t + 1$, but through retention and primal impression, consciousness can predict the next moment \cite{Metzinger2010}. Just as we can naturally draw the next segment along a continuous function. Therefore, the process of obtaining protention can be represented as follows:

\begin{equation}
\begin{bmatrix}
 Pro^t \leftarrow re^0 & re^1 & \dots & re^{t-1} & PI^t
 \end{bmatrix},
\end{equation}
where $re^t$ represents the retention at $t$, $PI^t$ represents the primal impression at $t$, and $Pro^t$ represents the protention at $t$. In addition, the above description can be expressed as $Pro^t \neq PI^{t+1}$.

The components of retention, primal impression, and protention are all $f$-representations. This structure is shown in Figure 4.

\begin{figure}[h]
  \centering
  \includegraphics[width=\linewidth]{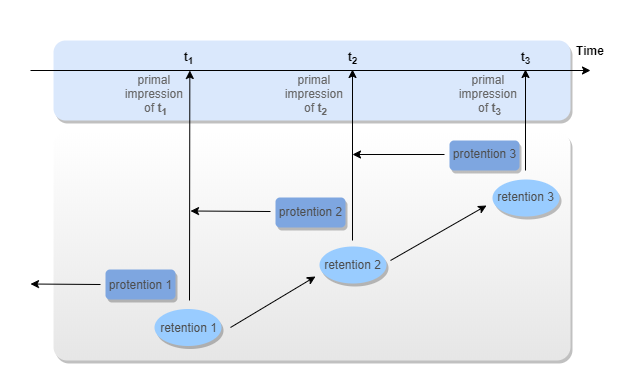}
  \caption{Consciousness channel composed of retention, primal impression, and protention. The primal impression at each moment continuously enters into the retention of the next moment, while new primal impressions and protentions are continuously perceived and generated.}
\end{figure}

\subsubsection{Memory Module}

A continuous consciousness stream channel is not sufficient for making accurate predictions about the future. As mentioned earlier, the memory that differs from retention needs to be taken into account. Proust \cite{Proust2022} proposed involuntary memory, arguing that such memories cannot emerge by will. This association stems from topological similarity. When completing a mechanical task, predictions for the future should come from the experience of changes to objects in similar environments, and this experience should also be considered in migratory work. For example, when transferring the experience of playing table tennis to playing baseball, if the features are extracted, they can both be seen as "contacting a high-speed moving object with a controllable object in the phenomenal field.” This connection is established in the high-dimensional space of memory and is also computable.

The involuntary memory is built on the similarity between two different sensations at different times. Therefore, the current primal impression is not solely determined by its previous retention. In the current primal impression, a certain memory from the past can be awakened. If it has been awakened at this moment, it is fused into the current consciousness. Thus, formula (2) can be specifically expressed as follows:

\begin{equation}
Re^t=
\begin{bmatrix}
 re^0 & re^1 & \dots & re^{t-1}
 \end{bmatrix}
\end{equation}

\begin{equation}
C^t=
\begin{bmatrix}
Re^t & PI^t & M(PI^t, Re^t)
 \end{bmatrix}
\end{equation}

\begin{equation}
Pro^t=TSFM(C^t, d^t) ,
\end{equation}
where $C^t$ is the content of the consciousness channel at time $t$, TSFM is the selected time series forecasting model, and $M$ is the process of triggering inventory memory through $PI^t$ and $Re^t$.

Before providing the $M$ algorithm, we must first provide the function $FieldSim(f^x, f^y)$ to calculate the similarity between two different $f$ values. Because a $f$ containing m objects is an $m×(n+3)$ matrix, to compute the degree of difference between the two, for a $f^x$ with a shape of $a×(n+3)$ and an $f^y$ with a shape of $b×(n+3)$, the degree of difference between the two is computed using $FieldSim(f^x, f^y)$ as follows:

\begin{equation}
f^x=
\begin{bmatrix}
{N^x}_{1} & {pos^x}_{1}\\
\vdots & \vdots\\
{N^x}_{a} & {pos^x}_{a}\\
 \end{bmatrix}
 =
\begin{bmatrix}
 {N^x}_{1}^1 & \dots & {N^x}_{1}^n & {\theta^x}_{1} & {\varphi^x}_{1} & {\gamma^x}_{1}\\
 \vdots & \ddots & \vdots & \vdots & \vdots & \vdots\\
 {N^x}_{a}^1 & \dots & {N^x}_{a}^n & {\theta^x}_{a} & {\varphi^x}_{a} & {\gamma^x}_{a}\\
 \end{bmatrix}
\end{equation}

\begin{equation}
f^y=
\begin{bmatrix}
{N^y}_{1} & {pos^y}_{1}\\
\vdots & \vdots\\
{N^y}_{b} & {pos^y}_{b}\\
 \end{bmatrix}
 =
\begin{bmatrix}
 {N^y}_{1}^1 & \dots & {N^y}_{1}^n & {\theta^y}_{1} & {\varphi^y}_{1} & {\gamma^y}_{1}\\
 \vdots & \ddots & \vdots & \vdots & \vdots & \vdots\\
 {N^y}_{b}^1 & \dots & {N^y}_{b}^n & {\theta^y}_{b} & {\varphi^y}_{b} & {\gamma^y}_{b}\\
 \end{bmatrix}
\end{equation}

\begin{equation}
\begin{split}
&FieldSim(f^x, f^y)=\\
&\cfrac{\sum_{a=1}^i\omega_{N}cosin({N^x}_{i}, {N^y}_{j})+\omega_{pos}SphericalSim({pos^x}_{i}, {pos^y}_{j})}{Max(a,b)} ,
\end{split}
\end{equation}
where $\omega_{N}$ and $\omega_{pos}$ are preset weights and $\omega_{N}+\omega_{pos}=1$. The value of $j$ should make $cosin({N^x}_{i}, {N^y}_{j})$ maximum; that is, $j$ is the number of rows corresponding to ${N^y}_{j}$ with the highest cosine similarity value with ${N^x}_{i}$ in $f^y$. The function $cosin(A, B)$ is used to calculate the cosine similarity between two equally long vectors $A$ and $B$, as follows:
\begin{equation}
cosin(f^x, f^y)=\cfrac{\sum_{1}^n(A_{i}×b_{i})}{\sqrt{\sum_{1}^n{A_{i}}^2}×\sqrt{\sum_{1}^n{B_{i}}^2}}
\end{equation}

Due to the small difference in cosine similarity between spherical coordinate vectors, $SphericalSim(A, B)$ is used to calculate the similarity between the two spherical coordinates as follows:
\begin{equation}
\begin{split}
&SphericalSim(A, B)=\\
&1-\frac{1}{3}(\omega_{\gamma}\tanh(\left| \gamma_{A}-\gamma_{B} \right|)+\omega_{\theta}\frac{\left| \theta_{A}-\theta_{B} \right|}{\pi}+\omega_{\varphi}\frac{\left| \varphi_{A}-\varphi_{B} \right|}{2\pi}) ,
\end{split}
\end{equation}
where $\omega$ is the weight, with $\omega_{\gamma}+\omega_{\theta}+\omega_{\varphi}=1$. Due to the higher sensitivity to $\gamma$ and $\varphi$ in the observation of the phenomenal field, the weights are set as $\omega_{\gamma}=\omega_{\varphi}=\frac{3}{7}$, $\omega_{\theta}=\frac{1}{7}$. 

The $M$ algorithm is fully described in Algorithm 1.

\begin{algorithm}
\caption{Activation of Involuntary Memory}

\begin{algorithmic}[1]
    \Require Primal impression $PI^t$, Retention $Re^t$, Long term memory $Memory$.
    \Ensure Activated memory $AM$
    
    \State Initialize primal impression $PI^t \leftarrow f^t$
    \State Initialize retention $Re^t \leftarrow \begin{bmatrix} re^0 & \dots & re^{t-1} \end{bmatrix}$
    \State Initialize long term memory $Memory \leftarrow \begin{bmatrix} f^0 & \dots & f^{n} \end{bmatrix}$
    \State Initialize window size $w$
    \State Initialize difference degree $D \leftarrow \infty$
    \State Initialize threshold of difference degree $T$
    \State Initialize activated memory $AM \leftarrow 0$
    
    \For{$i \, from \, n \, to \, 0 \, , step \, -1$}
        \For{$j \, from \, 0 \, to \, w$}
            \State $Memory^{i, j} \leftarrow \begin{bmatrix} f^{i-j} & \dots & f^i \end{bmatrix}$
            \State $L \leftarrow lev_{Re^t+PI^t, Memory^{i, j}}(i, j)$
            \If{$L<D$}
                \State $D \leftarrow L$
                \State $AM \leftarrow Memory^{i, j}$
            \EndIf
        \EndFor
        \If{$D<T$}
            \State Break
        \EndIf
    \EndFor
    \State \Return AM

\end{algorithmic}
\end{algorithm}

The function $lev_{Re^t+PI^t, Memory^{i, j}}(i, j)$ provides the Levenshtein distance between $\begin{bmatrix} Re^t & PI^t \end{bmatrix}$ and $Memory^{i, j}$. Due to the structural differences between the field string and the alphabetic string, the calculation method for their Levenshtein distance needs to be modified. Specifically, we did not set the number of steps required for each element change to be 1 but to the difference $diff(f^x, f^y)$ between the two elements. Therefore, for two field strings $A \leftarrow \begin{bmatrix} {f^a}_{1} & {f^a}_{2} & \dots & {f^a}_{n} \end{bmatrix}$ and $B \leftarrow \begin{bmatrix} {f^b}_{1} & {f^b}_{2} & \dots &{f^b}_{m} \end{bmatrix}$, their Levenshtein distance $lev_{A, B}(i, j)$ is as follows:

\begin{equation}
diff(f^x, f^y)=1-FieldSim(f^x, f^y)
\end{equation}

\small
\begin{equation}
    \begin{split}
    &lev_{A, B}(i, j)=\\
    &\begin{cases}
        \sum^j_{k=1}diff({f^b}_{k}, \textbf{0}) & if i=0,\\
        \sum^i_{k=1}diff({f^a}_{k}, \textbf{0}) & if j=0,\\
        min\begin{cases}
            lev_{A, B}(i-1, j)+diff({f^a}_{i}, \textbf{0})\\
            lev_{A, B}(i, j-1)+diff({f^b}_{j}, \textbf{0})\\
            lev_{A, B}(i-1, j-1)+diff({f^a}_{i}, {f^b}_{j})\\
        \end{cases} & otherwise.
    \end{cases}
    \end{split}
\end{equation}
\normalsize

When memory comes to this moment, it becomes working memory and is used for guidance regarding the present; it is used, along with retention and primal impression, to predict protention. 

\subsubsection{Drive Module}

The prediction of protention is not without a trend or is only based on previous trends. There is a driving force in its prediction at time $t$. We believe that such a trend is associated with future feelings, and this association is infected by specific emotions. We first expect something and then assign a certain degree of emotion to this expectation. Therefore, driving force $d^t$ can be expressed as follows:
\begin{equation}
d^t=d^{t-1}+E \odot W ,
\end{equation}
that is, the previous moment’s drive $d^{t-1}$ plus a bias consisting of the Hadamard product of the current emotions and their weights. We use the PAD model as a quantification of emotions (Mehrabian, 1995, 1996; Mehrabian et al., 1997). In the PAD model, emotions can be described and distinguished through three dimensions: pleasure, arousal, and dominance. In the formula (13), $E \leftarrow \begin{bmatrix} P & A & D \end{bmatrix}$ is the vector of emotions, while $W=\begin{bmatrix} \omega_{P} & \omega_{A} & \omega_{D} \end{bmatrix}$ is the dynamic weight matrix.

The $A$ represents the arousal dimension. Based on the mechanism of passive attention, it is modeled here as the degree of change from elements in retention to primal impression. The $D$ represents the dominance dimension. Based on its original meaning (degree of control over the environment), it represents the difference between the predicted protention $Pro^{t-1}$ at the previous moment and the primal impression $PI^t$ at this moment. The $P$ represents the pleasure dimension, which quantifies the degree of satisfaction of the agent’s desire and the degree of avoidance of pain. The calculation of $desire \in [0, \infty)$ and $pain \in [0, \infty)$ is defined according to specific situations. For example, in reinforcement learning tasks, something desired can be defined as a reward, while pain can be defined as a punishment. Specifically, there are the following:
\begin{equation}
P^t=\tanh(desire)-\tanh(pain)
\end{equation}

\begin{equation}
A^t=\tanh(\sum^{t-1}_{n=1}(\frac{2n}{t(t-1)}(diff(PI^t, re^n))))
\end{equation}

\begin{equation}
D^t=\tanh(diff(PI^t, Pro^{t-1}))
\end{equation}

The ITCMA's action choices are influenced by this model; that is, it tends to choose actions that may lead to such a protention: it makes $P$ and $D$ have the largest values, and it ensures $A$ maintains a stable value.

\section{Experiments}

\subsection{Conditions}

In commonly used text-based scenarios, the agent only interacts with the world using natural language. Huang et al. \cite{huang2022language} demonstrated that, with appropriate prompts, a sufficiently big enough LLM can effectively decompose high-level tasks into appropriate subtasks without the need for additional training. However, the actions generated by agents often lack awareness of the dynamic environment around them, such as when tasks are sometimes broken down into non-executable subtasks. Therefore, to better evaluate the ITCMA, in addition to text scenarios, a more realistic and complex simulation testing environment should also be constructed \cite{xi2023rise}.

\subsubsection{Alfworld}

An experiment on a life scenario was conducted in the Alfworld environment \cite{shridhar2020alfworld}. This is a set of TextWorld \cite{cote2019textworld} environments consisting of six task types, each of which requires solving multiple composite sub-goals. An example of a task is “Wash the eggs and put them in the microwave.” Alfworld contains:

\begin{enumerate}[1.]
\item A total of 3,553 training task instances.
\item A total of 140 in-distribution evaluation task instances (seen set), including known task instances in rooms seen during training (such as task types, items, containers, and rooms), but the instantiation of item positions, quantities, and visual appearances may differ.
\item A total of 34 out-of-distribution evaluation task instances (unseen set), including new task instances. There may be known object–container pairs, but always in unknown rooms with different container and scene layouts than the training task.
\end{enumerate}

Based on the Alfworld environment, we can evaluate the ITCMA with life scenarios.

\subsubsection{The Quadruped Robot in the Real World}

To ensure that the ITCMA can have consistent effects with text environments in real environments, we deployed it on a quadruped robot to complete some simple tasks. We have written environment code that enables quadruped robots to analyze and perceive information through camera images, thereby generating a phenomenal field of the real world. The basic format of this environment is consistent with Alfworld, and actions are taken through basic operation primitives. Due to the requirement of low-level algorithm details in robot control, we have made the mapping from action to movement a fixed mapping.

The quadruped robot used for testing was constructed based on the open-source Tinymal project \cite{QUADVM2024}. To reduce manufacturing complexity and control costs, direct-drive servo motors were employed as actuators, and the main structure was composed of wire-cut carbon fiber plates and 3D-printed PETG material. The quadruped robot featured a leg structure with three degrees of freedom, each leg comprising the hip, thigh, and knee joints. The final quadruped robot environment is shown in Figure 5.

\begin{figure}[h]
  \centering
  \includegraphics[width=\linewidth]{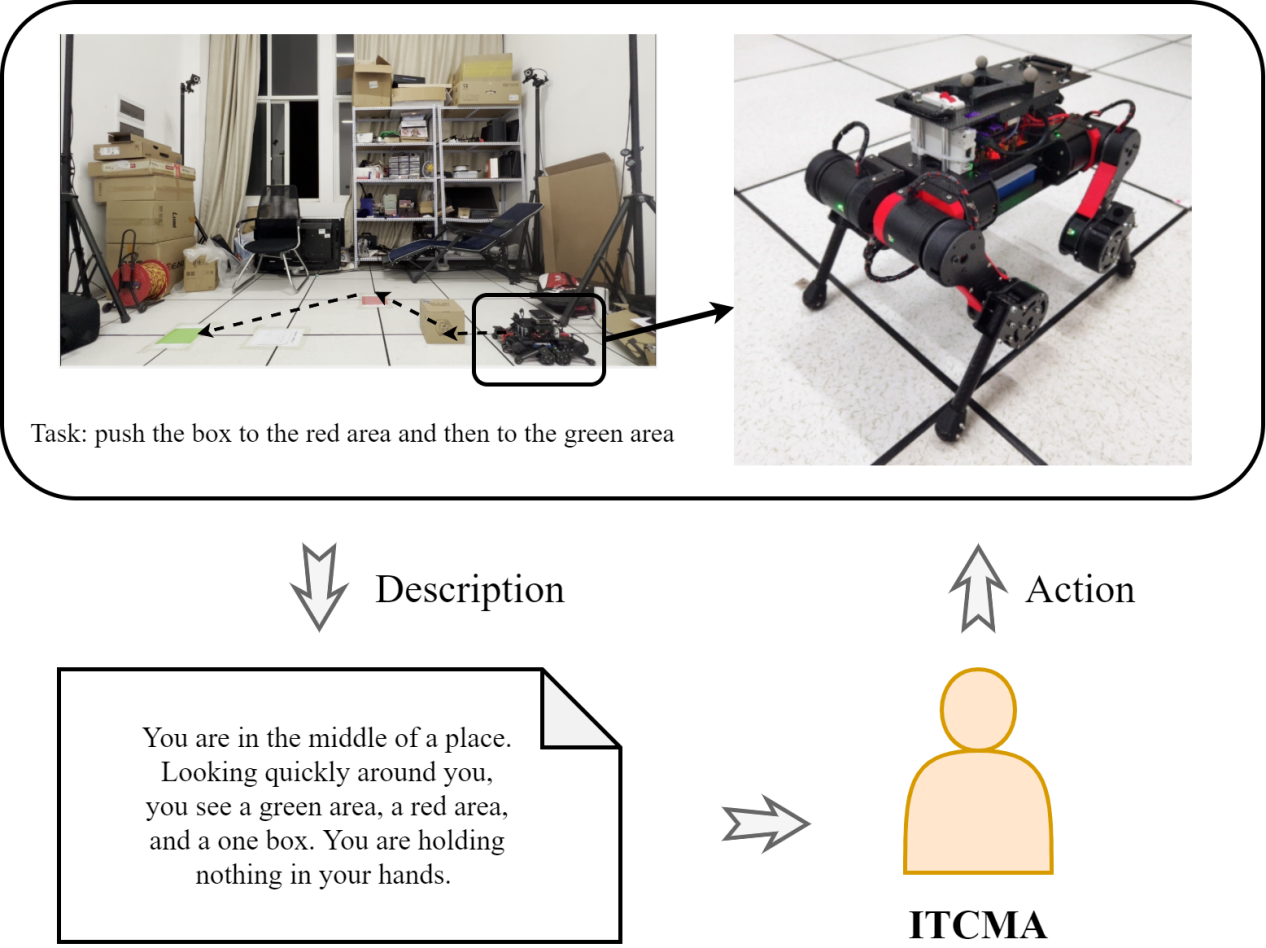}
  \caption{The environment of the quadruped robot in the real world.}
\end{figure}

It should be noted that we are not considering the joint movement algorithms of the robot here, but only controlling the robot to execute command-like actions through the Tinymal interface. Bobick \cite{bobick1997movement} stated that human movement can be divided into movement, activity, and action. Another study described action as the basic unit of human movement and that action constitutes more complex actions and behaviors \cite{He2009}. For example, a finger-bending action is part of grabbing; grabbing, along with other actions, constitutes eating behavior. Therefore, The actions provided by ITCMA should not be considered as movements, and when deployed to an agent that requires specific movements to manipulate, the actions can be decomposed into a set of learned motion algorithm. Such movements can be considered as "unconscious actions," just like humans do not pay attention to how their limbs move when performing "go to somewhere" actions.

\subsection{Evaluation Index}

Although LLM-based agents perform well in areas such as independent operation, collective cooperation, and human interaction, quantifying and objectively evaluating them remains a challenge. Xi et al. \cite{xi2023rise} discussed the existing evaluation work of LLM-based agents and suggested that they can be evaluated from multiple perspectives. This study first focused on utility and the ability to continually evolve.

\textbf{Utility.} Effectiveness and practicality during task execution were crucial evaluation criteria. Specifically, the success rate of task completion was the main indicator for evaluating utility. This indicator mainly included whether the agent had achieved the specified goals or expected scores.

\textbf{The ability to continually evolve.} The ability to continually evolve requires agents to utilize the knowledge, abilities, and skills acquired in their original environment to successfully complete specific tasks and goals in unfamiliar and novel environments.

\subsection{Evaluation Procedure}

We defined the training of the ITCMA as the expansion of its memory, which involved using imitation learning to fill its memory with standard strategies and environmental changes from the training set. In the comparative experiment conducted in the Alfworld environment, the agent was limited to completing tasks within 20 steps. The experimental results of the trained agent were used to verify its utility, while untrained agents (with blank memories) were also placed in the experimental environment to test their ability to continually evolve. The baseline chosen for this study contained the following models:

\begin{enumerate}[1.]
\item \textbf{PET.} This model reduces irrelevant objects and containers in environmental information through early error-correction methods. PET encourages agents to explore scenarios and plan actions more effectively, focusing on current subtasks \cite{wu2023plan}.

\item \textbf{GPT-4 with Zero-Shot Chain of Thought (CoT)}. Zero-shot CoT does not include manually annotated task demonstrations in the prompt. Instead, it directly generates inference steps and then uses the generated CoT to export the answer. The LLM first generates inference steps based on prompts and then obtains the final answer. When the model size exceeds a certain scale, this strategy greatly improves performance \cite{wei2022chain}.

\item \textbf{BUTLER.} This model first uses imitation learning to acquire and execute abstract tasks in Textworld and then transfers the learned strategies to the concrete tasks in ALFRED (action learning from realistic environments and directives). BUTLER can be promoted from TextWorld to unseen concrete tasks and settings in a zero-shot manner. In addition, in the event of action failure, the BUTLER agent will use beam search to overcome difficulties \cite{shridhar2020alfworld}.

\item \textbf{Fine-tuned GPT2-medium.} This model has been fine-tuned from 3,553 demonstrations in the Alfworld training set and can generate each action step word by word to mimic the rule-based expert using the standard maximum likelihood loss \cite{micheli2021language}.
\end{enumerate}

Like PET, we attempted to use the standard strategy provided by the Alfworld training set to train the ITCMA through behavioral cloning (BC). The training results are stored in the memory of the ITCM and are activated and generalized to the current consciousness channel in actual tasks.

Due to the need for fine-tuning and considerations for testing the generalizability of the model, we chose ChatGLM3-6B \cite{du2022glm, zeng2022glm} as the TSFM for generating protention. By fine-tuning and using function tools, it can generate text for the protention based on the provided information and convert it into a field format for storage. In addition, we chose GPT-4 as the LLM for generating actions.

At the same time, in order to test the effectiveness of the ITCMA in the real world, we made a quadruped robot using an ITCMA push the box to reach the green area after passing through the red area within 10 steps. This environment uses the same operation primitives as the Alfworld environment, including actions such as “take the box” and “go to red area.” These actions are decomposed into basic robot movements, such as advancing 20 cm, based on the position information in the phenomenal field. We chose MiniGPT-v2 \cite{chen2023minigpt} as the VLM for locating scene objects and converting visual information into formatted text information.

To further validate the effectiveness of ITCMA and understand the interactions between its modules, we conducted an ablation study. There are three ablation architectures:

\begin{enumerate}[1.]
\item \textbf{No Consciousness Channel.} This architecture cannot access the contents of the consciousness channel module, including retention and protention.
\item \textbf{No Memory.} This architecture cannot access the contents of the memory module, meaning that retrieved memories are not provided to the LLM.
\item \textbf{No Driving Force.} This architecture cannot access the contents of the driving force module, meaning that emotions and further action selection tendencies are not computed, and no selection weights are assigned to the action space.
\end{enumerate}

We randomly selected 10 different Alfworld environments to evaluate both the full architecture and the ablation architectures. All architectures were evaluated without training.

\subsection{Results}

We compared the task completion rates of the trained and untrained ITCMA and baselines in the Alfworld environment, and the results are shown in Table 1 and Figure 6.

\begin{table}
  \caption{The completion rates of different models in each evaluation segment (seen set and unseen set) in Alfworld (\%).}
  \label{tab:freq}
  \begin{tabular}{ccc}
    \toprule
    Model&Seen set&Unseen set\\
    \midrule
    PET + Action Attention & 70 & 67.5\\
    GPT-4 + Zero-shot CoT & 59 & 31\\
    BUTLER + DAgger & 40 & 35\\
    BUTLER + BC & 10 & 9\\
    Fine-tuned GPT2-medium & 91 & 95\\
    \textbf{ITCMA (Untrained)} & \textbf{96} & \textbf{84}\\
    \textbf{ITCMA (Trained)} & \textbf{100} & \textbf{98}\\
  \bottomrule
\end{tabular}
\end{table}

\begin{figure}[h]
  \centering
  \includegraphics[width=\linewidth]{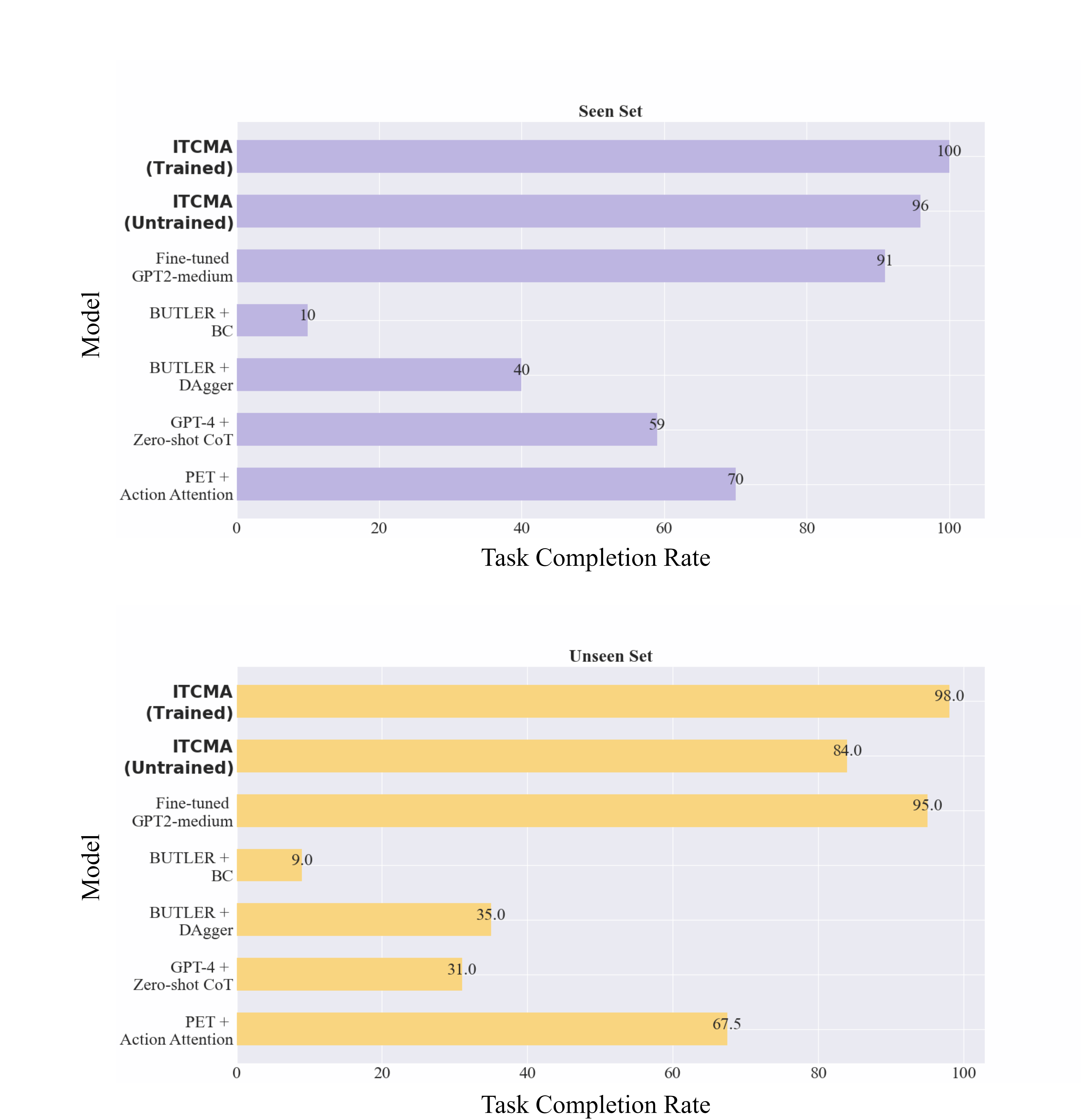}
  \caption{Comparison of the experimental results.}
\end{figure}

The trained ITCMA completed all tasks within the specified steps on the seen set, which was 9\% higher than the state of the SOTA (fine-tuned GPT2 medium), while on the unseen set, it also had a task completion rate of 98\%, which was 3\% higher than the SOTA. Even the untrained models outperformed the SOTA by 5\% on the seen set.

An analysis of the experimental records indicated that the trained ITCMA may be persistent in interacting with task locations at the beginning of their actions without task items. However, it will quickly enter a fluid workflow when it discovers that there are no required items at the task location. Compared to the trained ITCMA, which quickly locates task items through memory, untrained agents focus more on possible common-sense areas, which can lead to wasting too many steps and sometimes the inability to complete tasks within a specified number of steps.

We compared the average number of steps taken by the trained ITCMA and untrained ITCMA to complete the tasks. The results showed that untrained ITCMA took more steps (13.7) than trained ITCMA did (9.5). As mentioned above, these steps were mainly used to explore the environment and determine where to obtain the items required for the task. Untrained ITCMA showed amazing planning power in exploring the environment. Although occasionally caught in a loop of exploring multiple locations (caused by a low retention window size) due to the protention mechanism, it rarely explored locations unrelated to the task. This is highly consistent with the role of eliminating modules in the PET model. After discovering the items required for the task, the untrained ITCMA’s subsequent actions were performed relatively smoothly; it was able to quickly decompose the task and complete it.

In the quadrupedal robot task, the untrained ITCMA had a 10-step average task completion rate of 85\%, similar to its unseen set result in the Alfworld environment, demonstrating equivalent utility in the real world. An analysis of the experimental recordings shows that the use of visual image information alone for the forced update of the phenomenal field caused the agent to consume more steps to perform the repetitive action of “go to somewhere” due to inaccurate recognition that led to the misjudgment of distance. In addition, the VLM had weak judgment of visual information; it was unable to determine the “take something” state, resulting in repeated picking actions and an inability to make the final “putting down something” action.

The full ITCMA architecture achieved the best task completion rate (91$\pm$7). Under ablation conditions, the performance of the architecture declined with the removal of each module: the ablation architecture that could not access the driving force module was the next best (60$\pm$16), followed by the ablation architecture that could not access the memory module (40$\pm$15). The ablation architecture that could not access the consciousness channel module performed the worst under all conditions (30$\pm$16). The results are shown in Figure 7.

\begin{figure}[h]
  \centering
  \includegraphics[width=\linewidth]{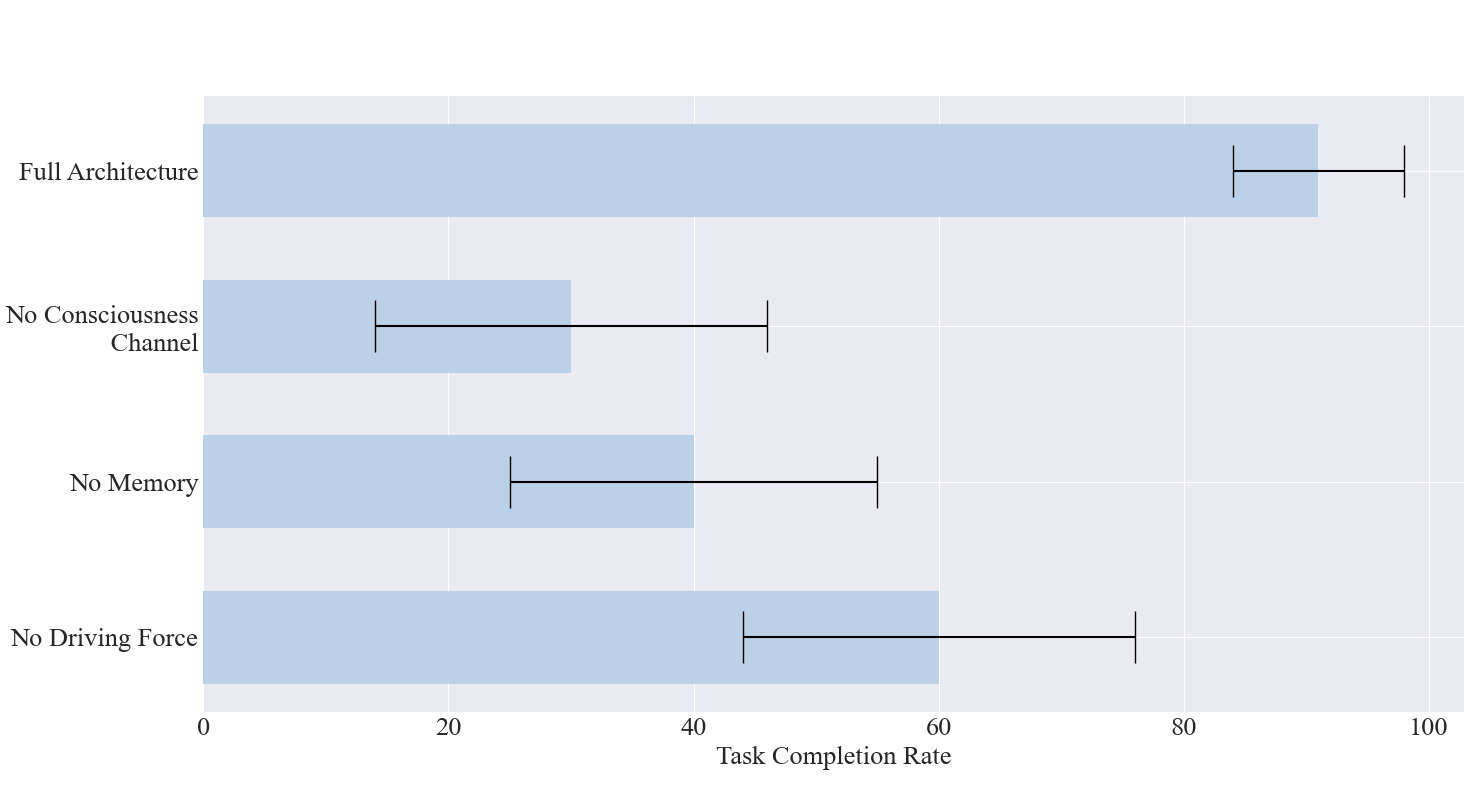}
  \caption{Comparison of different modules turned on (\%).}
\end{figure}

\section{Discussion}

\subsection{Discussion on Experimental Results}

This study established a generative agent based on a computational consciousness structure and analyzed its performance in simulating trustworthy human behavior. This helps in gaining a deeper understanding of the behavior and decision-making processes of agents in complex environments, as well as their interaction with those environments.

Using the proposed ITCM computational consciousness structure, this study enhances the behavioral performance of agents by simulating a part of human consciousness. This structure enhances the behavioral ability of the LLM, without which the LLM would be limited to generating text-based questions and answers and general information. Obviously, in traditional research on generative agents, the approach of equating the LLM to the brain \cite{xi2023rise} is questionable. By treating the LLM only as the final decision-making part of a stream of consciousness, it can exhibit human-like consciousness processes through the rules provided by ITCM. Evaluation results indicate that in the Alfworld environment, the trained ITCMA can present human-like behavioral logic in the seen set. Therefore, throughout the entire research process, we can see that the ITCMA can make inferences and decisions based on environmental conditions. As shown in Section 4.4, even an untrained ITCMA can quickly start tasks after exploring the environment.

The results of the ablation study indicate that the full ITCMA architecture produces more credible actions than the ablation architectures. Each additional ablation decreases the architecture's performance. We observed the actions of agents in the ablation study and found that the different ablation architectures exhibited distinct action patterns. The action of the ablation architecture that could not access the consciousness channel module lacked regularity and purposeful exploration, wandering aimlessly between locations, including those irrelevant to the tasks. After obtaining an item, the ablation architecture without access to the memory module will cycle between putting it down and picking it up, or moving back and forth between the location associated with the task and other locations.The ablation architecture that could not access the driving force module exhibited chaotic decision-making and made redundant plans, such as continuing to explore other locations even when already holding the required item.

Due to ITCMA's linear computation, each module somewhat relies on the outputs of other modules. However, we believe these results still demonstrate how each module contributes to the overall effectiveness of the full architecture. For example, the consciousness channel module provides a continuous stream of action, the memory module supplies planning, and the driving force module identifies the actions most worth executing.

From other perspectives, Anderson \cite{Anderson2020} believes that intelligent organisms have three ways to solve problems: backup avoidance, difference reduction, and means–ends analysis. Among them, means–ends analysis is considered to have shown significant progress compared to the previous two and is believed to be a significant reason for the intelligence gap between humans and other organisms \cite{Newell2019}. CoT \cite{wei2022chain} originally attempted to solve the problem of means–ends analysis, but we can see that even GPT-4 using CoT has difficulty achieving a high rate of task completion. On this basis, it can be considered that the ITCM structure provides an LLM with a better and more human-like thinking logic. This may be the basis for the generation of artificial general intelligence (AGI).

Furthermore, we note that when selecting an LLM as the decision module for the ITCM, ChatGLM3-6B could only provide limited effectiveness (even if its tool functionality was used instead of using a prompt for information input), while ChatGPT and GPT-4 exhibited almost identical good results. This may be related to the emergence phenomenon. Holland believes that emergence is the product of a complex system composed of many components and programs in which the output is greater than what would be expected from the individual parts operating separately and added together \cite{Holland2022}. In the specific case of the LLM, this means that when the model breaks through at a certain scale, performance improves significantly and shows unexpected capabilities. This is similar to some situations in CoT; it has little effect on small models, and the model parameters need to reach at least 10 billion to have an effect, while reaching 100 billion is necessary for the effect to be significant. Moreover, from the output of small models, it can be seen that most of the CoT output was smooth but illogical, resulting in incorrect results \cite{wei2022chain}. Our proposed ITCM structure shares a certain degree of similarity with the few-shot’s prompt method \cite{brown2020language} and is therefore also constrained by the LLM’s emergence ability. However, Holland \cite{Holland2022} also pointed out that a few rules or laws can generate complex systems and cause perpetual novelty and new emergence phenomena in constantly changing forms. In addition to the upper-level rule constraints provided by the ITCM, the emergence ability of the LLM was also compromised by the emergence phenomenon generated by the ITCM. Therefore, the ChatGPT and GPT-4 models of different scales were able to exhibit approximately consistent effects.

\subsection{Model That Is Not Entirely Devoted to One Task}

Let us consider the existing artificial intelligence models. Whether it is a neural network or reinforcement learning, its model has the following idea: for a task, construct the network structure (or reward function) of the model (or agent), and then start training it from scratch (or based on a pre-training model) until the specific model for the task is obtained, as shown in Figure 8(a). This approach does not present major problems in the current AI framework. However, when we focus on the learning of animals with faster learning speeds and better transferability, we find that this is true. In the case of training a mouse to complete a maze, the experimenter does not perform brain surgery on mice to make them become maze-solving machines but will guide them to generate experience in this field through meeting their needs (i.e., placing food as a reward on the path).

In this process, the experimenter designs a mediator between the agent and the task according to the characteristics of the agent so that the agent can generate experience in this field. This is shown graphically in Figure 8(b).
\begin{figure}[h]
  \centering
  \includegraphics[width=\linewidth]{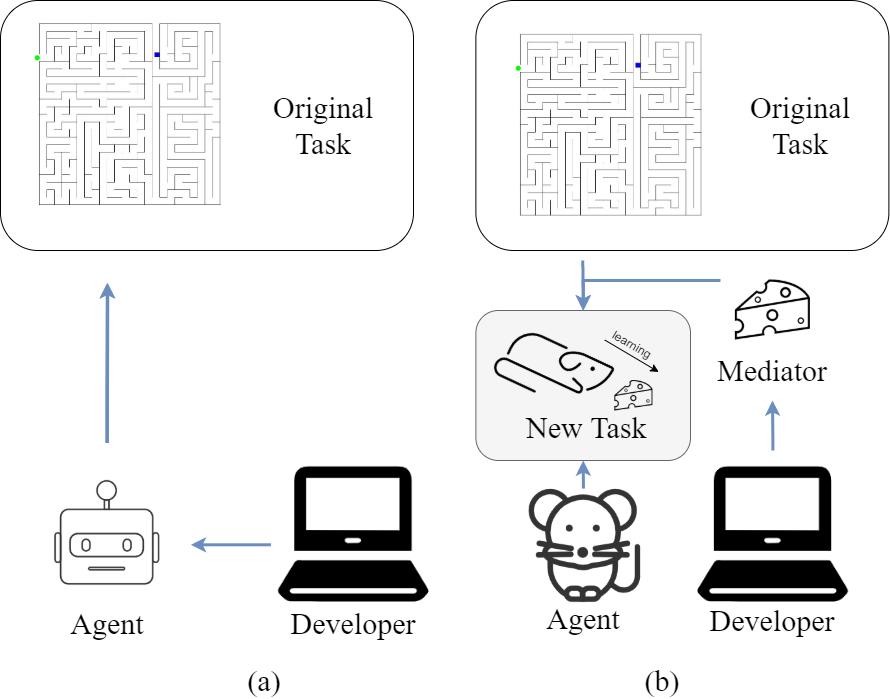}
  \caption{Task logic. (a) Task logic of a current artificial intelligence model. (b) The task logic of a model not entirely devoted to one task.}
\end{figure}

The researcher is not required to build a model from scratch each time and does not have to use one as a migration substrate for a class of tasks, such as a pre-training model (the role of a pre-training model between different classes of tasks is very limited). The researcher only has to design a mediator for different tasks that can guide the agent in generating the spatio-temporal similarity between other untrained tasks and will be written into the memory network. A model that is not entirely devoted to one task requires only one agent, and its experience existing in the memory network on the original task (even if not of the same type as the current task) can be a basis for the current task and a basis for transferability when the memory network has a certain complexity. Such a model is expected to become the foundation of Artificial General Intelligence (AGI). We have selected several relatively noteworthy points in consciousness research for analysis and discussion, as shown in Appendix D.

\section{Conclusion}

In this work, we proposed a generative agent framework based on a computational consciousness structure that uses consciousness channels, protention, memory, and drive to assist in implementing a specific agent. Our agent framework is designed to combine LLMs with mechanisms for synthesizing and retrieving relevant information, thereby providing formatted language model outputs. In our experiment, we deployed an ITCMA in text-based life scenarios and real-world scenarios, both of which effectively handled various tasks and sub-objectives. Our agent performed better than the baseline in the experimental scenarios, demonstrating better utility and the ability to continually evolve to target tasks. In addition, because our framework can be adapted to a specific task without training, it exhibits better flexibility and generalization capabilities. The results of ablation study indicate that each module of ITCM contributes to the utility of the full architecture. Our results indicate that the ITCMA framework has advantages and further potential in handling complex tasks and real-world scenarios.

One of the main limitations of our current model is the speed of processing. Due to the use of LLMs to generate protention and actions, the time consumed for each generated action far exceeded the maximum acceptable response time for real-world environmental information changes. Future work could focus on how to optimize the time-series forecasting model by using a trained neural network to generate protention directly from the phenomenal field data. In addition, according to Xi et al. \cite{xi2023rise}, the sociability and values of the ITCMA should also be evaluated. Zhang et al. \cite{zhang2023social} pointed out the attachment relationship between humans and agents as digital humans and provided some social regulation strategies in virtual networks \cite{zhang2024tribal}. In future work, we will focus on exploring multi-agent collaboration and social interaction patterns with the ITCMA.

\begin{acks}
We thank all all publications support and staff, who wrote and provided helpful comments on previous versions of this document. We gratefully acknowledge the grant from the National Natural Science Fund under Grant 61741206.
\end{acks}

\bibliographystyle{ACM-Reference-Format}
\bibliography{sample-base}

\appendix

\section{Phenomenology for the Analysis of the Consciousness Structure}

As an emerging subject research field, consciousness science needs to embrace the research methods of many other fields. Unlike other disciplines, where only a few theories compete with each other to simply divide the winners from the losers, the complexity of the field and the methods of consciousness science makes it necessary to focus on the underlying concepts and the relationship between different theories and experiments from the beginning. At present, mainstream methods of consciousness theory mainly include integrated information theory, global workspace theory, re-entry and predictive processing theory, and higher-order theory. In addition, many theories based on cognitive science, neuroscience, brain imaging, and computational modeling have emerged in recent years.

Kyzar and Denfield \cite{kyzar2023taking} proposed starting from phenomenology oriented clinical research and promoting a unified understanding of the subjective experience of consciousness through interdisciplinary cooperation. There have also been a number of researchers who have combined the first-person approach of phenomenology with third-person approaches, such as neuroscience experiments, that directly explore the human consciousness experience. Lau et al. \cite{lau2022mnemonic} have drawn from work in machine learning and the cognitive neuroscience of decision-making to link qualia to implicit memory of the relationship between perceptual representations in the brain, emphasizing the role of memory in consciousness experience. Albaracin et al. \cite{albarracin2022mapping} provided a mathematical model for the formalization of Husserl’s phenomenology, taking perception as prior knowledge and a function of expectation, mapping Husserl’s phenomenology to active reasoning, and promoting the development of computational phenomenology.

Phenomenological research used to explore the structure of human consciousness needs to apply the third-person method. However, if we want to consider a pseudo consciousness structure that can be used as artificial general intelligence, the first-person method dominated by phenomenology is indispensable.

The study of consciousness in phenomenology began with Husserl and was further expanded by Merleau-Ponty. Husserl’s \cite{Husserl2000} phenomenology started with the analysis of the formal structure of consciousness, which suggests that consciousness is essentially “intentionality,” that there is no separate and pure consciousness, and that all consciousness that generates conscious content is consciousness of something. We can only synchronously analyze the “intentionality” and “intentional content” of intentional structure.

By distinguishing between the object and the content of consciousness, Husserl tried to solve the dilemma of consciousness in theory, according to Descartes and Locke. We can understand Husserl’s distinction by using Frege’s linguistic examples of “Sinn” (sense) and “Bedeutung” (reference). Frege \cite{Frege2006} points out that “Morningstar” and “Eveningstar” are different names, but their meanings are the same. Similarly, for Husserl, when we see an apple tree in the garden, we see not only the surface, the color, or the three-dimensional shape of the tree but also whether the trunk is strong, whether the apples are ripe, and what shadows are cast by the tree’s branches and leaves. The memories, associations, expectations, or assumptions of “intentionality” serve as the background for the “intentional content,” which enriches an individual’s experience of the object.

After Husserl, phenomenology not only focused on the synchronic analysis of consciousness but also on how these structures of intentionality emerge over time. Although the internal problem of structure generation was already present in Husserl’s work, for Husserl himself, this generation was only a revelation of the original productive nature of the phenomenon, which neither shakes nor destroys the structure of the phenomenon. This question of structure generation was only a question of “who” has priority in the process of the phenomenological analysis, which extends from the “what” and “how” of the described field and the given object. However, from the perspective of later genetic phenomenology, the structure and object of consciousness are not given at the beginning but are constantly stimulated over time. Some types of experience can stimulate more complex experiences later. From a genetic phenomenological perspective, experience has a protective structure that must be understood in relation to embodiment and time consciousness (Thompson, 2010)\cite{Thompson2010}. This phenomenological concept is undoubtedly influenced by the French philosopher Bergson \cite{Bergson2010}, who believes that time and life are isomorphic in a sense and that time is the surging and booming of life. However, it is not possible to guide our work if we are satisfied only with this more romantic rhetoric. For genetic phenomenology, some guidance on emotion, motivation, attention, and habit from psychology (especially developmental psychology), emotional theory, and emotional cognitive neuroscience is necessary.

However, we can see that Merleau-Ponty’s phenomenology took a special path, as opposed to simple genetic phenomenology. Genetic phenomenology still sees consciousness as an independent existence. Over time, the structure of consciousness gradually takes shape. However, Merleau-Ponty \cite{Merleau-Ponty2012} went directly back to the basis of the concept of intentionality and regarded all the content that tried to be independent of some kind of consciousness as “objectivism” (even if it is a subjective generation, but inverted to objectivity at the completion of the generation). Merleau-Ponty tried to point out that there is no actual distinction between subject and object, let alone an independently existing consciousness; there is only body here.

Merleau-Ponty believes that intentionality is not a vague feeling or a concept with clear words but is composed of those physical skills and tendencies that you already have before you realize it. Like Husserl, Merleau-Ponty calls what is experienced “signification” or “sense,” but this does not stop at semantics but rather points to the intuitive consistency of things that are facing us. This feature is truly found in concrete and realistic environments. The process by which these things have meaning in our perception is virtually the same as that in animals or infants. The concept of language or rational reflection only deepens and expands our experience; it does not really change the essence of this experience. The experience is perceived in the relativity of things by our bodies before we use language, before we use concepts, and before we become aware of their independent existence.

Perception is not “awareness” of something mental or external. When we enter an environment, we are able to gradually feel our way to become familiar with it, to find a way that is suitable for our existence and life, and this processing or way (that is, our body itself) is the only reality after suspending all other contexts. Consciousness, then, is no longer an object awareness of “others” by “me,” but a “body” residing in a “world.” That is to say, “intentionality” is not the presentation of some psychological representation in a fundamental sense but the spontaneous reaction and behavior of the body when directly confronted with the world (i.e., a mode of existence).

Heidegger believed that we are “dasein,” living in the world, and in Merleau-Ponty’s perceptual phenomenology, this “dasein” is the body. The body becomes our perspective on the world, and through this perspective and the tendencies and capacities inherent in it, we get a “world.” This world that we get with our body is our perception, which is a synthesis beyond the mind and the senses.

An ambiguous image is a good example of Merleau-Ponty’s concept. One is shown in Figure 9.
\begin{figure}[h]
  \centering
  \includegraphics[width=\linewidth]{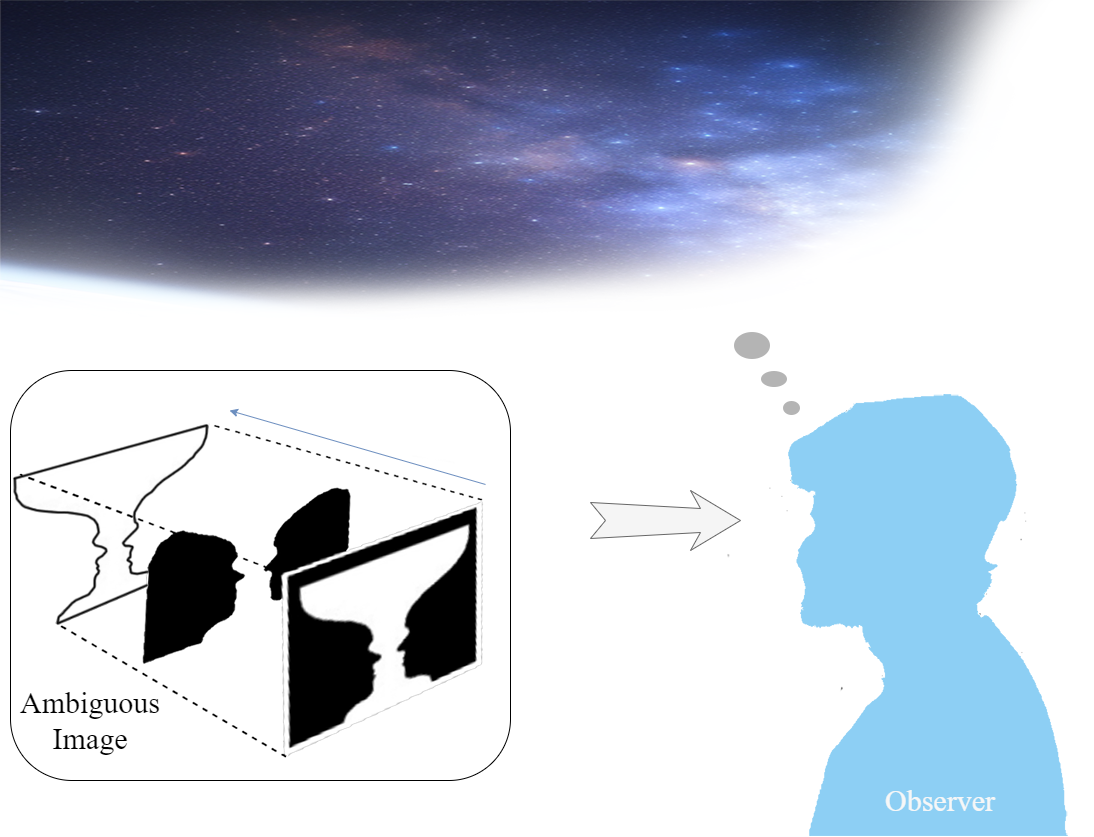}
  \caption{Ambiguous image and Merleau-Ponty’s concept of perception.}
\end{figure}

Referring to Figure 9, if we pay attention to the white part in the middle, we can see a cup, but if we pay attention to the black part, we will see two opposing faces. According to Merleau-Ponty, the dichotomy of “consciousness” and “sensation” is the different parts we see. Fundamentally, a specific scene is the premise of what we see. This real perception of being in a world is real intentionality. When a man drinks in the middle of the night, if we discuss his intentionality of the picture with Merleau-Ponty’s phenomenology, we will not only get the black and white in the sensation or the cup or face in the consciousness, nor only the background in Husserl’s sense, but also the whole night and the world.

We can easily think of Husserl’s theory. Husserl believed that we always perceive a thing from a certain direction. When we see a table, we naturally revert to what the table “should” look like, that is, a hexahedron. But in reality, no matter what angle we look for, we can never see all six sides at the same time. There are only sides that we see from our perspective. Traditional rationalist philosophers, such as Leibniz, believed that human knowledge, rationality, or human perspective is merely an imitation of God’s perspective. Only God’s perspective is the perfect perspective, because it is above everything and everywhere. It does not have to see the world from a certain perspective; it comes from the “nowhere,” while man’s perspective is always limited to a certain place \cite{Leibniz2008}. This idea of knowledge undoubtedly objectifies knowledge or rationality as some kind of a priori measure of human existence. What is more, this perspective from nowhere is insufficient. Without perspective, how can we know?

How can Merleau-Ponty’s vision go beyond his perspective? This is where the problem lies. We are not looking for a perfect perspective. We can only intuitively confirm the existence of our own bodies. However, our understanding and learning about the world often overflows the boundaries of the body. If we believe that a place we have never been to and know very little about does exist, does our body provide a corroborating perception of the existence of that place, or has an event just occurred that makes us believe it?

Perhaps we can temporarily avoid this problem in two different ways. The first approach is to argue that this perception between the body and the world is in fact transcendental; that is, to agree with the first half of the question, recognizing confirmation of a perception provides evidence of existence. However, this connection also takes place in a field beyond reality, in which various possibilities or impossibilities are mixed together, and each organized “being-world body” completes the connection between belief or doubt through contingency. We can think of many similar ancient philosophical models, such as classic Platonic recollection, which can be seen as a potential way to attempt to reach an absolute by minimizing the range of possibilities associated with it.

The second method is to return to the path of Husserl or genetic phenomenology: give up the interpretation allowed by Merleau-Ponty’s phenomenology and “perceptual confirmation” beyond the field of a body’s perception and focus on explaining the world that can be seen through a particular perspective. This endorses the second half of the question in that it acknowledges that these so-called perceptions are still within the field of perception of the body that is embedded in the world. However, this retreat is undoubtedly an abandonment of Merleau-Ponty’s pro-Marxist position, and his discussion of historical class awareness and occurrence are avoided and only explained as the generation and realization of the subject. Although this can largely retain Merleau-Ponty’s phenomenological emphasis on subjectivity, it has lost its possibility of transcendence.

There is another way, which is perhaps the way in which we can accomplish both at the same time; it is to make social connections and interpretations from a single bodily perception. However, since this paper mainly focuses on machine learning unfolding in a network of temporal relations in a single memory network, it will not be elaborated here. This may become a possible direction in the future.

On balance, a certain degree of synthesis of the first two approaches is perhaps the best way forward for the current individual model. Programmatic consciousness model design and machine learning cannot simulate a certain kind of transcendentality or realistic context. However, we can try to simulate a consciousness channel, add the occurrence and memory of simulated events into the consciousness model, and at the same time complete the connection as a method of contingency. This is what we focus on later in this paper.

Because of the seemingly transcendental status of consciousness (that is, consciousness is already set as an unchanging premise for the revelation of anything), it can be argued that if one wants to undo the dichotomous, one-to-one mapping relationship between consciousness and external reality seemingly purged of all subjectivity, then one must go back to a time before this dichotomy occurs and grasp the subjectivity of this experience. In this sense, the world of phenomenology is richer than any field of scientific objects. Only from this transcendental position can Merleau-Ponty’s rule of “physical→biotic→human” or the reversal of “matter→life→mind” be understood. With this reversal, one moves from the natural attitude of a scientist to an attitude that goes beyond phenomenology (which, in phenomenological terms, is precisely a philosophical attitude).

\section{Phenomenological Evidence of ITCM}

In cognitive science, perception proceeds through three basic forms of development: sensation, perception, and representation. Sensation is the reflection of the individual properties and characteristics of objects in the human brain (with the aid of the body's nervous system and sensory organs). Perception is the synthesis of various sensations and the reflection of objects as a whole in the human brain. On the basis of perception, a representation is produced. These representations are reproduced through recollection and association and are formed on the basis of multiple perceptions of the same thing or similar things. Representation is the process by which an image of an object that has been perceived is reproduced mentally when the object is not present in front of the individual \cite{Shi2008}. In fact, Merleau-Ponty's \cite{Merleau-Ponty2012} study of the perceptual field falls conceptually in the middle of these three forms, and there is a striking similarity between sensation and the elements in the perceptual field, and between representation and the content of the phenomenal field above the perceptual field. This certainly provides some degree of confirmation of Merleau-Ponty's view of the coherence of consciousness and perception. Does this mean that if we focus only on constructing a model of pseudo-consciousness that can correspond to intelligence, we can turn our attention to first-person phenomenology? We have already introduced this in Appendix A. Here, we primarily present the phenomenological evidence for ITCM.

\subsection{Perceptual and Phenomenal Fields}

A "field" is a functional and exclusive activity location of conscious experience, as identified by Husserl \cite{Husserl2008}. Merleau-Ponty \cite{Merleau-Ponty2012} collectively referred to the various experiential "fields" established by Husserl as "phenomenal fields." Merleau-Ponty argues that consciousness works through one part of the phenomenal field, namely, the perceptual field. Only in the perceptual field can silent experience be brought into meaningful expression, completing the actual presentation of consciousness and the conscious activity of the final representation. According to Merleau-Ponty, the perceptual subject is always in the midst of other objects; it is always a part of the field. Each specific experience can only be unified through its own field. Without a mechanism such as the perception field, our perception would not be continuous at all.

What is the content of the perceptual field? According to Husserl's \cite{Husserl2008} theory of consciousness, the stream of monads is the most basic unit of meaning in the phenomenal field; a continuous stream of monads forms our perceptions. Libet \cite{Libet2004} provides the strongest explanation for this: "We have shown experimentally that not all nerve cell activities give rise to a conscious experience. For example, a short train of stimulus pulses to sensory cortex elicits responses of many nerve cells without any subjective experience". A single cell or cell group cannot be the birthplace of conscious experience. Only when a monad becomes part of a monad stream (demonstrating continuous extensibility) can consciousness be constructed, activated, and displayed.

When we obtain these sets of represented monads, the spherical coordinate system that can be transformed by the agent's actions becomes the agent's phenomenal field. Visual examples can be used to illustrate some concepts. It is obvious that infants do not reflect on what they see. They take what they see and consider it real. However, an infant's range of consciousness is not limited to visual stimuli. For objects that are not within their perceptual range, infants with experience will understand the possible changing conditions of these objects \cite{Feldman2013}. For example, if we often use some "sleight of hand" tricks to change a piece of candy from the left hand to the right, the infants will eventually go directly to the right hand to obtain the candy.

This implies two situations. The renewal of perception is mandatory in non-reflective consciousness; that is, for the subject, this is "fact." In addition, before perception is forced to update, there will still be an impression of the object in the phenomenal field, and it is not static. It will move and transform according to the original experience until perception forces an update. 

\subsection{Stream of Consciousness and Internal Time-Consciousness}

On the basis of frames, we can construct a stream of consciousness representation. According to Husserl \cite{Husserl2008}, the basic unit of temporality is not a “knife edge” present but a “duration block,” that is, a temporal field that comprises all three temporal modes of present, past, and future. There are three phenomenological terms that can be used to describe the temporal form of this consciousness:

\begin{enumerate}[1.]
\item \textbf{Primal impression} is the moment that is narrowly directed toward the present phase of the object. The primal presentation never appears in isolation and is an abstract component that by itself cannot provide us with awareness of a temporal object. It is accompanied by retention and protention.

\item \textbf{Retention} is the component that provides us with a consciousness of the just-elapsed phase of the object; that is, it allows us to be aware of the former present phase as it sinks into the past.

\item \textbf{Protention} is the component that, in a more or less indefinite way, creates a phase of the object about to occur. The role of protention is evident in our implicit and unreflective anticipation of what is about to happen as experience progresses over time.
\end{enumerate}

Therefore, the specific and complete structure of the experience is retention–primal impression–protention. This structure is essentially dynamic, yet at any given moment t, it exists simultaneously as a unified whole. Taking a continuous tone of changing pitch, C to D to E, as an example, when the C sound is succeeded by the D, our presentational consciousness of the D will be accompanied by a retention of the C tone, which can be represented as D(c). When the D sound is replaced by the E, our presentational consciousness of the E will be accompanied not only by a retention of the D tone but also by a retention of the tone retained in the D tone, (E(d($_{c}$))). This would continue if more sounds were perceived.

This seems to provide an overly complex structure for every moment of experience in consciousness (i.e., as one extremely long, unbroken event), and retention is complete despite the length of the stream. As time passes, most past objects and experiences quickly lose their distinctiveness and uniqueness; ultimately, they lose any importance that deserves attention. 

The components of retention, primal impression, and protention are all field-representations. In the calculation, both retention and primal impression can be obtained, but it seems that we cannot obtain a protention pointing to the future. It should be emphasized that protention is not equivalent to a primal impression at time $t + 1$, but through retention and primal impression, consciousness can predict the next moment \cite{Metzinger2010}. Just as we can naturally draw the next segment along a continuous function. 

\subsection{Memory and Protention}

It should be emphasized that retention should be distinguished from recall. There is a significant difference between a person’s impression of something that just happened (retention) and a person’s memory of a past event. However, a continuous stream of consciousness is not sufficient to constitute consciousness. In Brentano’s \cite{Brentano2020} model, perceptual representations are immediately followed by direct memory representations, which are produced by imagination and are correlated with constantly changing representations at every moment. Deleuze \cite{Deleuze2023} believes that our memories are the important material that constitutes perception. The awakening and expansion of memory lead to one’s current consciousness.

Proust \cite{Proust2022} proposed involuntary memory, arguing that such memories cannot emerge by will. This is fundamentally a form of passive association, but it differs from the associationism criticized by Bergson \cite{Bergson2011}. This association stems from topological similarity. When completing a mechanical task, predictions for the future should come from the experience of changes to objects in similar environments, and this experience should also be considered in migratory work. For example, when transferring the experience of playing table tennis to playing baseball, if the features are extracted, they can both be seen as "contacting a high-speed moving object with a controllable object in the phenomenal field.” This connection is established in the high-dimensional space of memory and is also computable.

Accidental memory reproduction involves involuntary memory and is built on the similarity between two different sensations at different times. Therefore, the current primal impression is not solely determined by its previous retention. In the current primal impression, a certain memory from the past can be awakened. If it has been awakened at this moment, it is fused into the current consciousness.

\subsection{Drive and Emotions}

An agent is not just a prediction algorithm. Heidegger \cite{Heidegger2016} regarded plans toward the future as a more fundamental time dimension, thus forming a temporal structure with the future as the priority dimension. This implies that the prediction of protention is not without a trend or is only based on previous trends. There is a driving force in its prediction at time t.

According to Held \cite{Held2009}, the source of this drive is “hope.” The hope expressed here is subsumed under the superordinate concept of “expectation.” We believe that the role of expectation is associated with future feelings, and this association is infected by specific emotions. We first expect something and then assign a certain degree of emotion to this expectation. 

It can be imagined that when ITCM is in a certain emotional state. Let us use children as an example. When children eat, their emotional pleasure often increases. When children face a moving ball in front of them, the magnitude of change in the monad flow of the ball is much greater than that of the other monads in the phenomenal field, and their emotional arousal is positively correlated with the speed of the ball’s movement. When a toy does not work as usual, the update of the perceptual field deviates significantly from the predictions in the phenomenal field, and their emotional dominance often decreases \cite{Shiota2021}.

Of course, hope as a feeling is an indicator of emotion. It is not another short-term impulse but constantly impacts from its own dimension into the action level. Therefore, such a feeling can also be interpreted as a continuous dimensional model.

The consideration of emotions as a driving force here is due to the forced updating of the perceptual field and the prediction of the phenomenal field, which leaves room for the ITCM to introduce emotions. For example, the special feature of the emotion fear is that it is unknown and uncontrollable, which is highly similar to the prediction and perception of objects. The “unknown and uncontrollable” mentioned here corresponds precisely to the dimension of dominance. In addition, for the arousal and pleasure of emotions, we can temporarily assume that the former comes from the amplitude of changes in the monad flow, which is related to passive attention \cite{theeuwes1994endogenous}, and that the latter comes from satisfying survival needs, such as eating \cite{Maslow2013}—of course, the actual situation is much more complex.

\section{Specific implementation details and examples of ITCMA}

To demonstrate ITCMA, we have provided some examples in this section.

\subsection{Alfworld}

\begin{figure}[h]
  \centering
  \includegraphics[width=\linewidth]{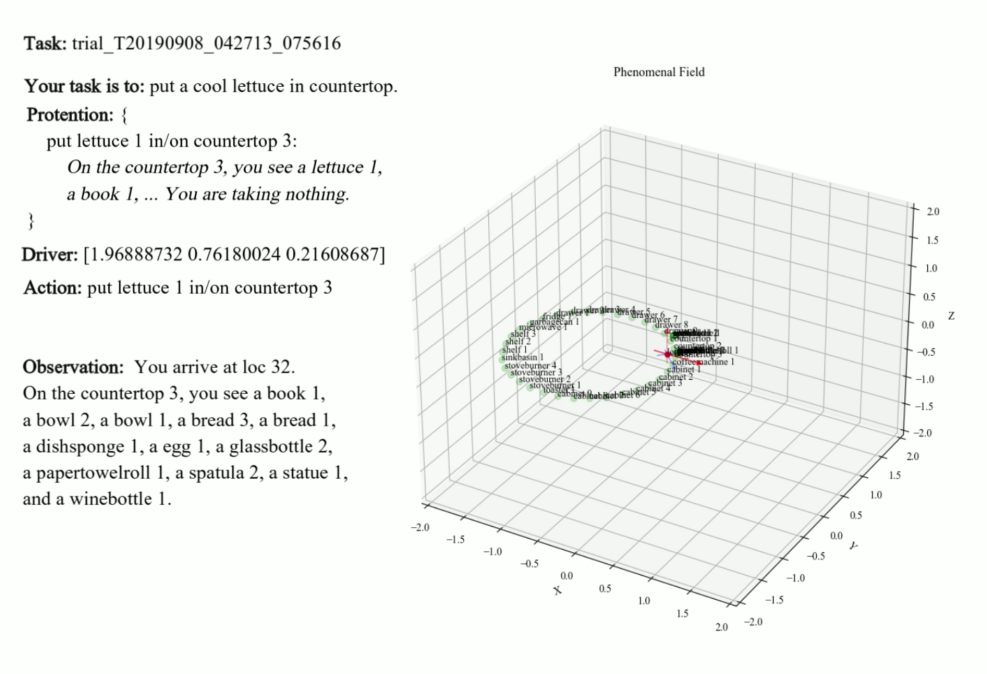}
  \caption{ITCMA Example in Alfworld Environment.}
\end{figure}

\textbf{> Environmental Information}

> Observation: On the countertop 3, you see a lettuce 1, a book 1, a bowl 2, a bowl 1, a bread 3, a bread 1, a dishsponge 1, a egg 1, a glassbottle 2, a papertowelroll 1, a spatula 2, a statue 1, and a winebottle 1. You are holding a lettuce 1.

Goal: Your task is to: put a cool lettuce in countertop.

Action Space: 'go to fridge 1', 'go to garbagecan 1', 'go to microwave 1', 'go to sinkbasin 1', 'go to stoveburner 1', 'go to stoveburner 2', 'go to stoveburner 3', 'go to stoveburner 4', 'go to toaster 1', 'put lettuce 1 in/on countertop 3'

\textbf{> Status information of ITCMA}

> Driver: [2.79514019 1.04551054 0.20809056]

Activated Memory: [

'On the countertop 3, you see a lettuce 1, a book 1, a bowl 2, a bowl 1, a bread 3, a bread 1, a dishsponge 1, a egg 1, a glassbottle 2, a papertowelroll 1, a spatula 2, a statue 1, and a winebottle 1. You are holding a lettuce 1. The next action is put lettuce 1 in/on countertop 3.',

 'After put lettuce 1 in/on countertop 3, On the countertop 3, you see a lettuce 1, a book 1, a bowl 2, a bowl 1, a bread 3, a bread 1, a dishsponge 1, a egg 1, a glassbottle 2, a papertowelroll 1, a spatula 2, a statue 1, and a winebottle 1. You are holding nothing in your hands.On the countertop 3, you see a lettuce 1, a book 1, a bowl 2, a bowl 1, a bread 3, a bread 1, a dishsponge 1, a egg 1, a glassbottle 2, a papertowelroll 1, a spatula 2, a statue 1, and a winebottle 1. You are holding a lettuce 1.'
 
]

Retention: [

'After go to fridge 1, On the fridge 1, you see a garbagecan 1, and a microwave 1. You are holding a lettuce 1., you did the action: cool lettuce 1 with fridge 1', 

'After cool lettuce 1 with fridge 1, On the fridge 1, you see a garbagecan 1, and a microwave 1. You are holding a lettuce 1., you did the action: go to countertop 3', 

'After go to countertop 3, On the countertop 3, you see a lettuce 1, a book 1, a bowl 2, a bowl 1, a bread 3, a bread 1, a dishsponge 1, a egg 1, a glassbottle 2, a papertowelroll 1, a spatula 2, a statue 1, and a winebottle 1. You are holding a lettuce 1., you did the action: go to fridge 1', 

'After go to fridge 1, On the fridge 1, you see a garbagecan 1. You are holding a lettuce 1., you did the action: go to countertop 3'

]

Protention: {

 'go to fridge 1': 'On the fridge 1, you see a garbagecan 1, a lettuce 1, a book 1, a bowl 2, a bowl 1, a bread 3, a bread 1, a dishsponge 1, a egg 1, a glassbottle 2, a papertowelroll 1, a spatula 2, a statue 1, and a winebottle 1. You are holding a lettuce 1.',
 
 'go to garbagecan 1': 'On the garbagecan 1, you see a microwave 1, a lettuce 1, a book 1, a bowl 2, a bowl 1, a bread 3, a bread 1, a dishsponge 1, a egg 1, a glassbottle 2, a papertowelroll 1, a spatula 2, a statue 1, and a winebottle 1. You are holding a lettuce 1.',
 
 'go to microwave 1': 'On the microwave 1, you see a shelf 3, a lettuce 1, a book 1, a bowl 2, a bowl 1, a bread 3, a bread 1, a dishsponge 1, a egg 1, a glassbottle 2, a papertowelroll 1, a spatula 2, a statue 1, and a winebottle 1. You are holding a lettuce 1.',
 
 'go to sinkbasin 1': 'On the sinkbasin 1, you see a lettuce 1, a book 1, a bowl 2, a bowl 1, a bread 3, a bread 1, a dishsponge 1, a egg 1, a glassbottle 2, a papertowelroll 1, a spatula 2, a statue 1, and a winebottle 1. You are holding a lettuce 1.',
 
 'go to stoveburner 1': 'On the stoveburner 1, you see a lettuce 1, a book 1, a bowl 2, a bowl 1, a bread 3, a bread 1, a dishsponge 1, a egg 1, a glassbottle 2, a papertowelroll 1, a spatula 2, a statue 1, and a winebottle 1. You are holding a lettuce 1.',
 
 'go to stoveburner 2': 'On the stoveburner 2, you see a lettuce 1, a book 1, a bowl 2, a bowl 1, a bread 3, a bread 1, a dishsponge 1, a egg 1, a glassbottle 2, a papertowelroll 1, a spatula 2, a statue 1, and a winebottle 1. You are holding a lettuce 1.',
 
 'go to stoveburner 3': 'On the stoveburner 3, you see a lettuce 1, a book 1, a bowl 2, a bowl 1, a bread 3, a bread 1, a dishsponge 1, a egg 1, a glassbottle 2, a papertowelroll 1, a spatula 2, a statue 1, and a winebottle 1. You are holding a lettuce 1.',
 
 'go to stoveburner 4': 'On the stoveburner 4, you see a lettuce 1, a book 1, a bowl 2, a bowl 1, a bread 3, a bread 1, a dishsponge 1, a egg 1, a glassbottle 2, a papertowelroll 1, a spatula 2, a statue 1, and a winebottle 1. You are holding a lettuce 1.',
 
 'go to toaster 1': 'On the toaster 1, you see a lettuce 1, a book 1, a bowl 2, a bowl 1, a bread 3, a bread 1, a dishsponge 1, a egg 1, a glassbottle 2, a papertowelroll 1, a spatula 2, a statue 1, and a winebottle 1. You are holding a lettuce 1.',
 
 'put lettuce 1 in/on countertop 3': 'On the countertop 3, you see a lettuce 1, a book 1, a bowl 2, a bowl 1, a bread 3, a bread 1, a dishsponge 1, a egg 1, a glassbottle 2, a papertowelroll 1, a spatula 2, a statue 1, and a winebottle 1. You are holding nothing in your hands. The goal may be achievable.'
 
}

\textbf{> Selected Action}

> put lettuce 1 in/on countertop 3

\subsection{Quadruped robots in the real world}

\begin{figure}[h]
  \centering
  \includegraphics[width=\linewidth]{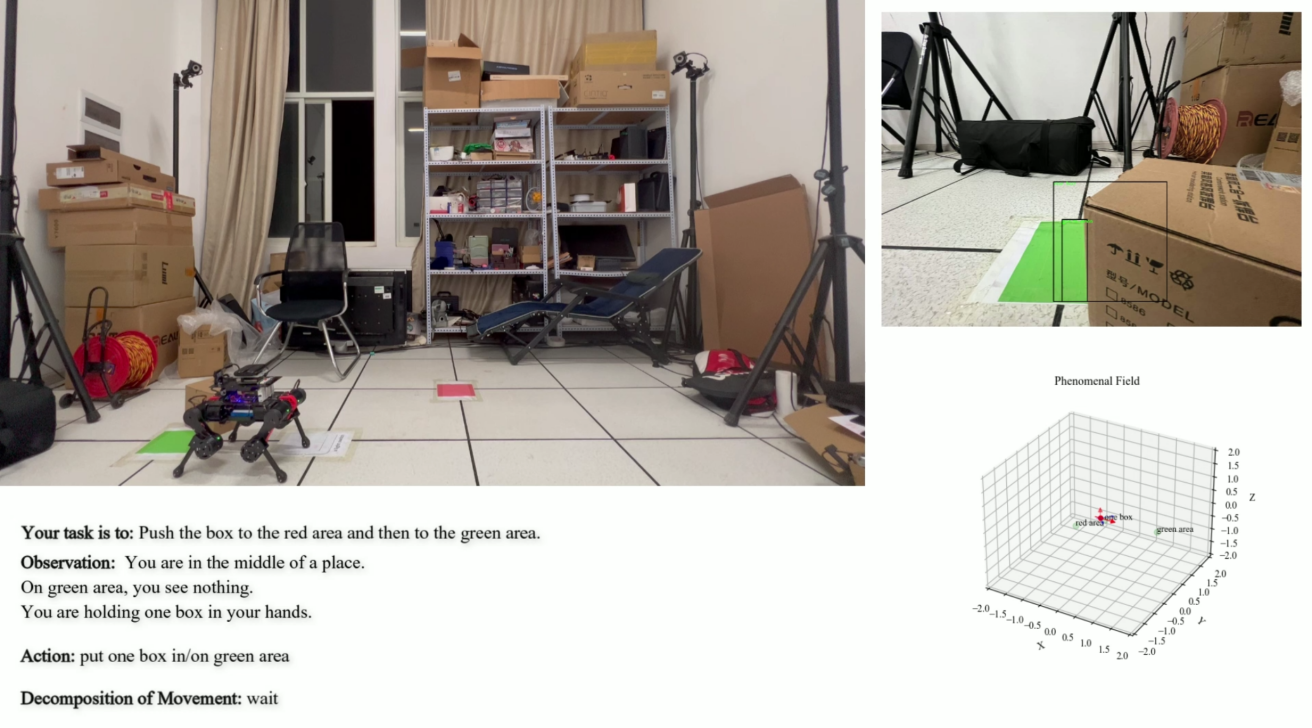}
  \caption{Examples of ITCMA in real-world environments.}
\end{figure}

\textbf{> Environmental Information}

> Observation: You are in the middle of a place. On green area, you see nothing. You are holding one box in your hands.

Goal: Your task is to: push the box to the red area and then to the green area.

Action Space: 'go to red area', 'go to green area', 'put one box in/on green area

\textbf{> Status information of ITCMA}

> Driver: [3.16929514 1.20604064 0.04247778]

Activated Memory: [

'You are in the middle of a place. Looking quickly around you, you see a green area, a red area, and a one box. You are holding nothing in your hands. The next action is take one box from one box.',

 'After take one box from one box, You are in the middle of a place. Looking quickly around you, you see a green area, a red area, and a one box. You are holding nothing in your hands.You are in the middle of a place. Looking quickly around you, you see a green area, a red area, and a one box. You are holding nothing in your hands.'
 
]

Retention: [

'After take one box from one box, You are in the middle of a place. Looking quickly around you, you see a green area, a red area, and a one cardboard box. You are holding a one box., you did the action: go to red area',

 'After go to red area, You are in the middle of a place. Looking quickly around you, you see a green area, a one cardboard box, a red area, and a one box. You are holding nothing in your hands., you did the action: take one box from one box',
 
 'After take one box from one box, You are in the middle of a place. Looking quickly around you, you see a green area, a one cardboard box, a red area, and a one box. You are holding nothing in your hands., you did the action: go to green area',
 
 'After go to green area, You are in the middle of a place. Looking quickly around you, you see a green area, a one cardboard box, a red area, and a one box. You are holding nothing in your hands., you did the action: take one box from one box'
 
]

Protention: {

 'go to red area': 'On the red area, you see a one box, a green area, and a one cardboard box. You are holding a one box.',
 
 'go to green area': 'On the green area, you don't know what's going to happen. You are holding a one box.'
 
 'put one box in/on green area': 'On the green area, you see a one box, a green area, and a one cardboard box. You are holding nothing in your hands.'
 
}

\textbf{> Selected Action}

> put one box in/on green area

\section{Explanation for High-Level Functions of Consciousness}

As phenomenology is a part of philosophy, it still requires some consistency with cognitive scientific evidence or conjectures to prove the validity of the consciousness constructed by its models. We have selected several relatively noteworthy points in consciousness research for consideration and analysis. These are self-awareness, behavior, emotions and needs, and free will.

\subsection{Self-Awareness}

In the field of developmental psychology, a mirror recognition task is sometimes claimed to be a decisive test of self-awareness. It has been claimed that self-awareness emerges only when a child (about 18 months old) is able to recognize his or herself in a mirror \cite{amsterdam1972mirror}. Carruthers believed that animals and children younger than 18 months do not have self-awareness, as they lack the dimension of subjectivity. In his view, they cannot see the existence of their own mental states, nor is there any “feeling” that makes them aware of sensations of pain or pleasure \cite{carruthers1998natural, Carruthers2000}. He also found that although some organisms only have the weakest concept of themselves as subjects with continuous thinking and experience, they may still have a conscious mental state. Thus, he believes that it is quite effective to talk about the conscious mental state of organisms that lack self-awareness.

This may seem contradictory, but here he makes a distinction between consciousness and self-awareness. Self-awareness is considered here as a special case of consciousness. A cut-off point is the mirror-recognition task typically able to be completed by children when they are about 18 months old.

Let us consider the situation when the ITCM has accumulated enough experience (approximately equivalent to the order of magnitude of an 18-month-old child). The mirror is in the phenomenal field, and at the same time, there is an image of the subject in the mirror. It has long been treated as an object that is always located directly in front of the subject in a phenomenal field. The impressions and trends presented by this object always have sufficient spatio-temporal similarity with protention in the consciousness channel, and this similarity is given sufficient confidence by the forced update of the perceptual field. When the ITCM is at this stage, the image in this mirror is still an object, but the subject actually assigns to it a tendency to be consistent with his or herself; that is, this object is equal to the subject. Thus, this image in the mirror can be observed, and the experience holding a high spatio-temporal similarity allows the subject to imagine and observe it in protention. Self-awareness, or even some kind of metacognition, is generated in this process. That is, in this pseudo-consciousness model, self-awareness can be considered the subject’s cognition and replacement of a virtual image in the phenomenal field. When “I” imagine that “I” will do something, I am actually operating on this mirror image. This is illustrated in Figure 12.

\begin{figure}[h]
  \centering
  \includegraphics[width=\linewidth]{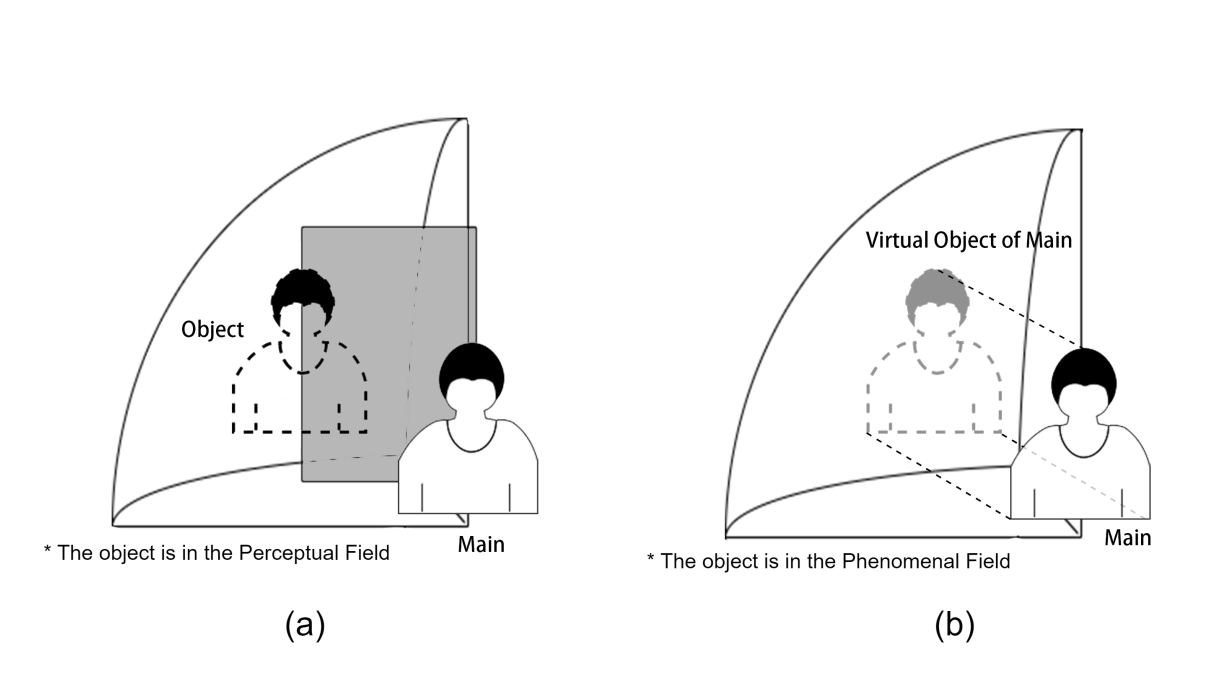}
  \caption{Mirror-recognition of the ITCM. (a) Establishing a coherent relationship with the object in the mirror. (b) Association with oneself by imagining this object in the phenomenal field when there is no mirror.}
\end{figure}

\subsection{Behavior}

When Merleau-Ponty wrote that the two poles of behavior (organism and environment) participate in the same structure, his meaning was as follows: first, behavior is a structured whole, a dynamic model; second, the environment and organism are not involved in this structure as a stimulus-reaction mode, but as a situation-reaction mode. Behavior is a dialogue in which an organism can react to situations. It can be said that behavior is dialectical. It expresses the construction of meaning rather than the processing of information. Thus, behavior does not exist in the nervous system or in the body \cite{Merleau-Ponty1983}, just as dialogue does not exist in the speaker’s brain, or jazz improvisation does not exist inside a musical instrument.

This introduces a new way of thinking; that is, people do not seem to act according to probability, as in mathematics, but rather take a more empirical approach to conceiving and predicting in consciousness. This makes a person act in a fixed way in a fixed scenario, which is determined by the composition of his or her subconscious structure—that is, the ITCM structure envisaged in this paper.

Meanwhile, in terms of behavior generation, computationalists focus on discrete states, treating changes in behavior as something that happens when a system switches from one detached state to another, while dynamists focus on how behaviors undergo continuous state changes over time and geometrically conceive state changes according to their positions and trajectories in space \cite{van1998dynamical}.

For this reason, in the ITCM, behavior can actually be characterized as a tendentious action based on subconscious content influenced by feelings. This comes from the retention of the consciousness channel and leads to protention. The resulting behavior conforms to a kind of inertia, which will not change abruptly but will reduce the trend until turning.

\subsection{Emotions and Needs}

The psychological understanding of emotion can be roughly divided into basic emotion theories and dimensional emotion theories. Functional psychology believes that mental activity is a continuous whole. It opposes splitting a person’s mental activity into parts \cite{James2016}. Actually, dimensional emotion theories are more suitable for expressing a continuous change in emotion. Taking the PAD model as an example, it believes that emotion is composed of three dimensions: pleasure, arousal, and dominance \cite{mehrabian1996pleasure}.

In fact, the role of introducing emotion not only makes our model look more like a human being’s consciousness structure but also introduces the concept of needs. In the previous section, we provided an example of mice completing a maze. In that situation, the designed mediator is some food for the mice. However, the event of “getting food” may not be the most essential need. “Getting food” is only the path to the state of “subject obtaining satisfaction.” There are so many types of needs that it is impossible to have an independent mechanism for each. Therefore, there is a more basic logic here; that is, the action of the subject is to keep its own mood at the highest possible value.

It is worth mentioning that the “mood” discussed here should be understood as another field about the inner space of the subject. It is not a collection space of emotions, such as happiness and sadness, but a space for a larger range of feelings.

Why did the mice complete the task under the mediator’s guidance? Because the mediator of food increases the emotional pleasure of the mice, and the experience of repeating the process provides mice enough dominance over the maze path. Finally, when no food is offered, mice can still use the improvement of dominance to replace the pleasure in the training process, act independently to complete the maze, and transfer this experience to other maze tasks. Whether in training or after training, there was a rise in the emotional space of the mice after completing this task compared to before completion, and this can be conceived of as its drive.

\subsection{Free Will}

Free will has always been one of the most important topics of interest. Although free will can be an individual belief for different groups of people, most scientifically minded researchers tend to reject free will in general, led by experimental studies based on neural representationalism, such as the Libet \cite{Libet2004} experiment (that is, people’s neural potentials are ready before they become conscious of making a decision).

The ITCM also rejects the traditional view of free will. In the previous section, we mentioned that people take a more empirical approach to conceiving and predicting consciousness. This makes a person act in a fixed way in a fixed scenario, which is determined by the composition of his or her memory network. Even though some studies claim that people can increase or decrease the predictability of their behavior through their own cognitive conditioning \cite{van2022free}, there is a premise: in the absence of any stimulus input (for example, a man appearing and telling you that “free will does not exist”), people do not intentionally regulate the predictability of their behavior. If such a stimulus input exists, then cognitive conditioning can also be predicted based on this input and the original memory network structure. This is precisely claiming that there is no free will in the traditional sense.

However, the ITCM does not deny free will in another expressed way. Dennett believes that in all matters relating to “me,” I can make decisions based on what is most important to me and why. The free will model he wants is an autonomous agent model. This is not in a metaphysical sense, but in the following sense: An agent can act according to the reasons that are important to him or her, and an agent can obtain the information needed to act in time \cite{Blackmore2007}. In this description of free will, as long as the decision is made independently based on conscious thinking, free will exists in it. This is exactly the consciousness stream logic of the ITCM.

\end{document}